\documentclass[]{gewu}

\usepackage{wasysym}

\usepackage[toc,page,header]{appendix}

\usepackage{minitoc}
\usepackage{microtype}      
\usepackage[dvipsnames,table]{xcolor}         

\usepackage{hyperref}
\usepackage{url}
\usepackage{booktabs}
\usepackage{multirow}
\usepackage{subcaption}
\usepackage{caption}
\usepackage{amssymb}
\usepackage{graphicx}
\usepackage{amsmath}
\usepackage{wrapfig}
\usepackage{enumitem}
\usepackage{arydshln} 
\usepackage{gensymb}
\usepackage{makecell}
\usepackage{pifont}

\newcommand{\modalityicon}[1]{%
    \raisebox{-0.15ex}{\includegraphics[height=0.85em]{#1}}%
}

\title{ROMA: LLM System for Real-World Object-Centric Multi-Sensory Active Perception}

\author[1,2,3,*]{Ruoxuan Feng}
\author[4,*]{Yutong Chen}
\author[1,2]{Ruihua Song}
\author[3]{Huan Yang}
\author[3]{Zhongyuan Wang}
\author[3,5]{Guocai Yao}
\author[1,2,3,6,\textnormal{\Letter}]{Di Hu}

\affiliation[1]{Gaoling School of Artificial Intelligence, Renmin University of China}
\affiliation[2]{Beijing Key Laboratory of Research on Large Models and Intelligent Governance}
\affiliation[3]{Beijing Academy of Artificial Intelligence}
\affiliation[4]{Beijing Jiaotong University}
\affiliation[5]{State Key Laboratory of Multimedia Information Processing, Peking University}
\affiliation[6]{AresoX}

\contribution[*]{ Equal Contribution}
\contribution[\textnormal{\Letter}]{ Corresponding author}

\abstract{
Humans inherently understand the physical world through an active process. When sensory evidence is insufficient to infer physical properties, we naturally interact with the environment by deciding \textit{what information is missing}, \textit{how to acquire it}, and \textit{when sufficient evidence has been obtained}. In stark contrast, existing multi-sensory robot systems mainly integrate sensory inputs rather than actively acquiring missing evidence through interactions. In this work, we introduce \textbf{ROMA}, an LLM-based system for \textbf{R}eal-World \textbf{O}bject-Centric \textbf{M}ulti-Sensory \textbf{A}ctive Perception. ROMA integrates vision, audio, tactile, and force sensing into a \textit{reasoning-interaction-feedback} loop. The model identifies missing evidence and determines the target objects, interactions, and perception modalities, while a physical interface executes the selected interactions and collects the multi-sensory feedback. To support this capability, we construct \textbf{ROMI-2K}, a large-scale real-world multi-sensory object interaction dataset covering nearly 2,000 objects and 6 atomic interactions with synchronized multi-sensory feedback. 
Building on these data, we develop a two-stage training framework that first aligns sensory modalities and then equips the LLM to assess evidence sufficiency, select informative interactions, and reason over the multi-sensory feedback.
We further characterize active perception as \textit{perception chains}, where acquired multi-sensory evidence guides subsequent interactions and reasoning, and establish \textbf{ROMA Bench} to evaluate single-attribute, long-horizon multi-attribute, and intent-driven active perception.
Experiments show that ROMA can actively acquire missing evidence and solve complex, long-chain multi-sensory perception tasks that many existing methods struggle to handle, laying a strong perceptual foundation for active multi-sensory embodied agents.
}

\MyEmail{Ruoxuan Feng at \email{fengruoxuan@ruc.edu.com}, Di Hu at \email{dihu@ruc.edu.com}}

\checkdata[Project Page]{\url{https://gewu-lab.github.io/ROMA/}}
\checkdata[Checkpoint \& Dataset]{\url{https://huggingface.co/datasets/GeWuLab/ROMA}}
\checkdata[Code]{\url{https://github.com/GeWu-Lab/ROMA}}

\begin{document}
\maketitle


\begin{figure}[t]
    \centering
    \includegraphics[width=0.98\linewidth]{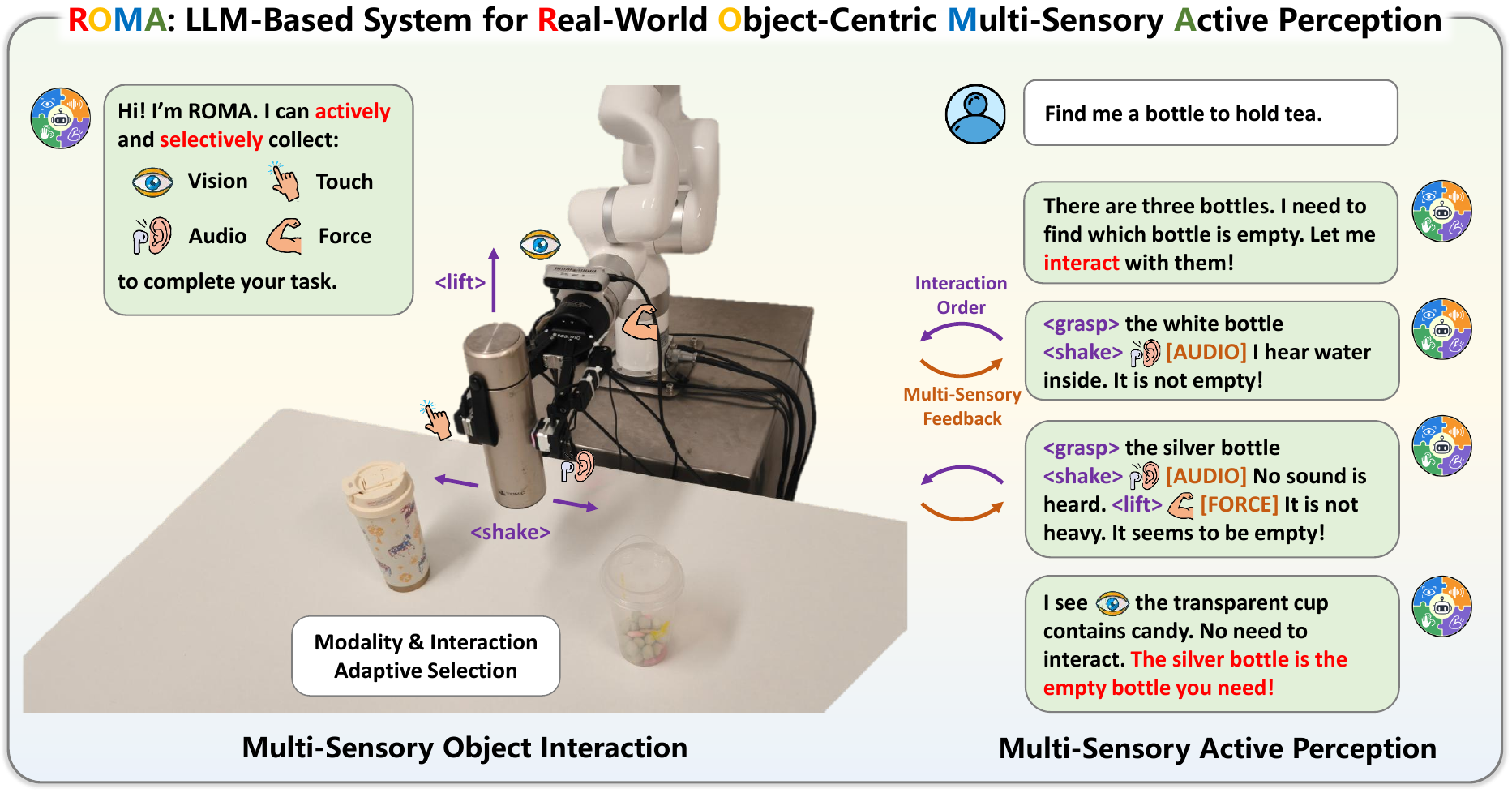}
    \caption{\textbf{ROMA} is an LLM-based system for Real-World Object-Centric Multi-Sensory Active Perception. Through a physical interface, ROMA actively acquires missing sensory evidence by integrating
\modalityicon{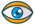} visual,
\modalityicon{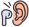} audio,
\modalityicon{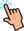} tactile, and
\modalityicon{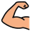} force feedback in a \textit{reasoning-interaction-feedback} loop. Given an instruction, ROMA can identify the information required for the task and selectively determine the target object, interaction, and sensory modality to acquire the necessary evidence.}
    \label{fig:teaser}
\end{figure}

\section{Introduction}

\textit{I saw. I touched. I understood.} This is how humans actively build an understanding of the physical world. When vision alone is insufficient to infer an object's physical properties, we naturally interact with it to acquire additional sensory evidence. We lift an object to infer its weight through force feedback, or shake it to investigate its contents through audio cues.
As different interactions provide different sensory feedback and reveal different object properties, effective multi-sensory perception requires more than simply integrating available sensory inputs. It also requires actively determining what evidence is missing, how to acquire it, and when sufficient evidence has been obtained.

Recent advances in multi-modal foundation models~\citep{team2026qwen35} and robotic sensors~\citep{lambeta2024digitizing} have made it increasingly feasible to equip robots with both broad reasoning capabilities and diverse perception modalities. 
However, most existing multi-sensory embodied systems still treat sensory observations as given inputs rather than evidence to be actively acquired~\citep{li2023see,feng2025play,liu2025mla}. They mainly focus on modality integration, while largely overlooking the active evidence acquisition through physical interaction. 
Recent efforts have begun to move beyond this passive paradigm. 
MOSAIC~\citep{tatiya2024mosaic} studies interactive multi-sensory object property learning but exhaustively performs pre-defined interactions. MultiPLY~\citep{hong2024multiply} enables an LLM to select physical interactions for multi-sensory reasoning but remains limited to simulation with simplified interactions and sensory feedback.
Learning real-world active perception requires large-scale data covering diverse objects, interactions, sensory feedback, and annotations, yet existing multi-sensory resources remain limited in both scale and interaction supervision.

To address these challenges, we establish a real-world active perception framework spanning both data and system. We first construct \textbf{ROMI-2K}, a large-scale real-world multi-sensory object interaction dataset covering nearly 2,000 everyday objects and 6 atomic interactions. It provides synchronized visual, audio, tactile, and force feedback. We combine a modified multi-sensory UMI~\citep{chi2024universal} for efficient single-object collection with a multi-sensory robotic arm for real tabletop scenes. This setup systematically captures diverse object responses across individual objects and multi-object environments.
Beyond raw sensory feedback, we additionally annotate object identities, physical properties, contents, and interaction-specific attributes. The annotations cover diverse properties such as hardness, material and inside contents. We further construct scene-level active perception tasks from these annotations. These tasks provide diverse combinations of physical attributes, interactions, and sensory feedback, enabling systematic study of real-world multi-sensory active perception.

Building on these resources, we propose \textbf{ROMA}, an LLM-based system for \textbf{R}eal-World \textbf{O}bject-Centric \textbf{M}ulti-Sensory \textbf{A}ctive Perception. 
ROMA integrates vision, audio, tactile, and force sensing into a \textit{reasoning-interaction-feedback} loop. It consists of a ROMA-7B LLM that actively determines what information to acquire and a physical interface that executes the selected interactions and collects sensory feedback. 
At the model level, ROMA enables the Multi-Sensory LLM to identify missing information and determine how to acquire it. It adaptively selects the target object, interaction, and modalities at each step, and recognizes when sufficient evidence has been obtained. This enables the model to actively acquire information rather than passively reasoning over fixed sensory inputs.
At the interaction level, the physical interface translates the decisions into executable actions through a grasp interface and 6 atomic interactions, supported by a training-free grasping policy~\citep{fang2023anygrasp} and dedicated grasp filtering and refinement mechanisms.
It then collects the required multi-sensory feedback and returns it to the LLM for further reasoning.
Together, these components enable ROMA to close the active perception loop in the real world, continuously acquiring task-relevant evidence and adapting its perception process based on sensory feedback.

To evaluate this active capability, we characterize active perception as \textit{perception chains}, where sequential interactions and sensory feedback are selected and acquired based on prior evidence to determine a target object property. We then establish \textbf{ROMA Bench} with three task types: single-chain tasks targeting a single attribute, multi-chain tasks requiring the perception of multiple attributes, and intent-driven tasks that infer and perceive relevant attributes from implicit user intent. We evaluate ROMA and existing MLLMs on both offline benchmark and online real-world scenarios, including the latest GPT-6 Astra. Results show that many existing MLLMs struggle to localize target objects, select appropriate interactions, and acquire the multi-sensory evidence required for complex long-horizon and real-world active perception tasks. In contrast, ROMA can continuously interact with the environment, acquire complementary sensory evidence, and progressively build a richer understanding of objects and their physical properties. We hope this work paves the way for a new generation of active multi-sensory embodied agents.

\section{Related Works}
\subsection{Active Perception} 
Active perception~\citep{bajcsy1988active} studies how an agent actively acquires additional information when its current observations are insufficient. Existing work primarily selects informative viewpoints or locations to reduce visual uncertainty~\citep{chaplot2021seal,xiong2025vision,liu2026activevla,hong2026esi,liu2026sapave}. Recent studies further explore physical interactions for acquiring tactile feedback that reveals local surface properties~\citep{shahidzadeh2024actexplore,schneider2025tactile,schneider2026apple}. MultiPLY~\citep{hong2024multiply} extends active perception to audio, vision, tactile, and temperature modalities with an LLM, but remains limited to simulated environments with simplified interactions and sensory feedback. In contrast, we enable active multi-sensory perception in real-world settings from both the data and system perspectives.

\subsection{Multi-Sensory Embodied AI}
 Recent embodied AI systems increasingly integrate multiple sensory modalities to improve perception and interaction with the physical world~\citep{liu2025mla,jones2025beyond}. Among them, tactile sensing captures contact-rich properties such as texture and deformation~\citep{feng2025anytouch,higuera2025sparsh}, audio reveals object materials and interaction events~\citep{wang2025sound,liu2025maniwav}, and force feedback characterizes interaction dynamics~\citep{he2025foar,yu2026forcevla}. Recent works have also explored the complementary roles of multiple modalities during manipulation~\citep{feng2025play,li2023see,chen2025multi,tatiya2024mosaic}, alongside large-scale real-world multi-sensory data collection~\citep{gao2022objectfolder,clarke2025x}. However, existing systems primarily use heterogeneous sensory inputs passively, while ROMA actively selects interactions and sensory modalities to acquire task-relevant feedback.

\subsection{Physical Reasoning}
Recent studies investigate physical reasoning across vision, audio, and tactile modalities to infer object properties and physical behaviors. Vision primarily supports reasoning about materials and spatial relationships~\citep{gao2024physically,chen2024spatialvlm}, while audio and tactile sensing provide complementary cues about materials, contents, contact events, hardness, and surface properties~\citep{yu2022pacs,zong2025counterfactual,yu2024octopi,xie2026universal,lyu2026tacreasoner,zong2026vitatouch,tu2026unitac}. Recent works further combine multiple modalities to exploit their complementary information~\citep{hong2024multiply,lai2026touch}. 
However, existing methods mainly perform physical reasoning over passively acquired sensory observations, without actively acquiring missing physical evidence in real-world environments. ROMA addresses this limitation by enabling real-world active physical reasoning through object interactions and multi-sensory feedback.

\section{ROMI-2K: Object-Centric Multi-Sensory Interaction Dataset}
\label{sec:dataset}

\begin{figure}[t]
    \centering
    \includegraphics[width=0.99\linewidth]{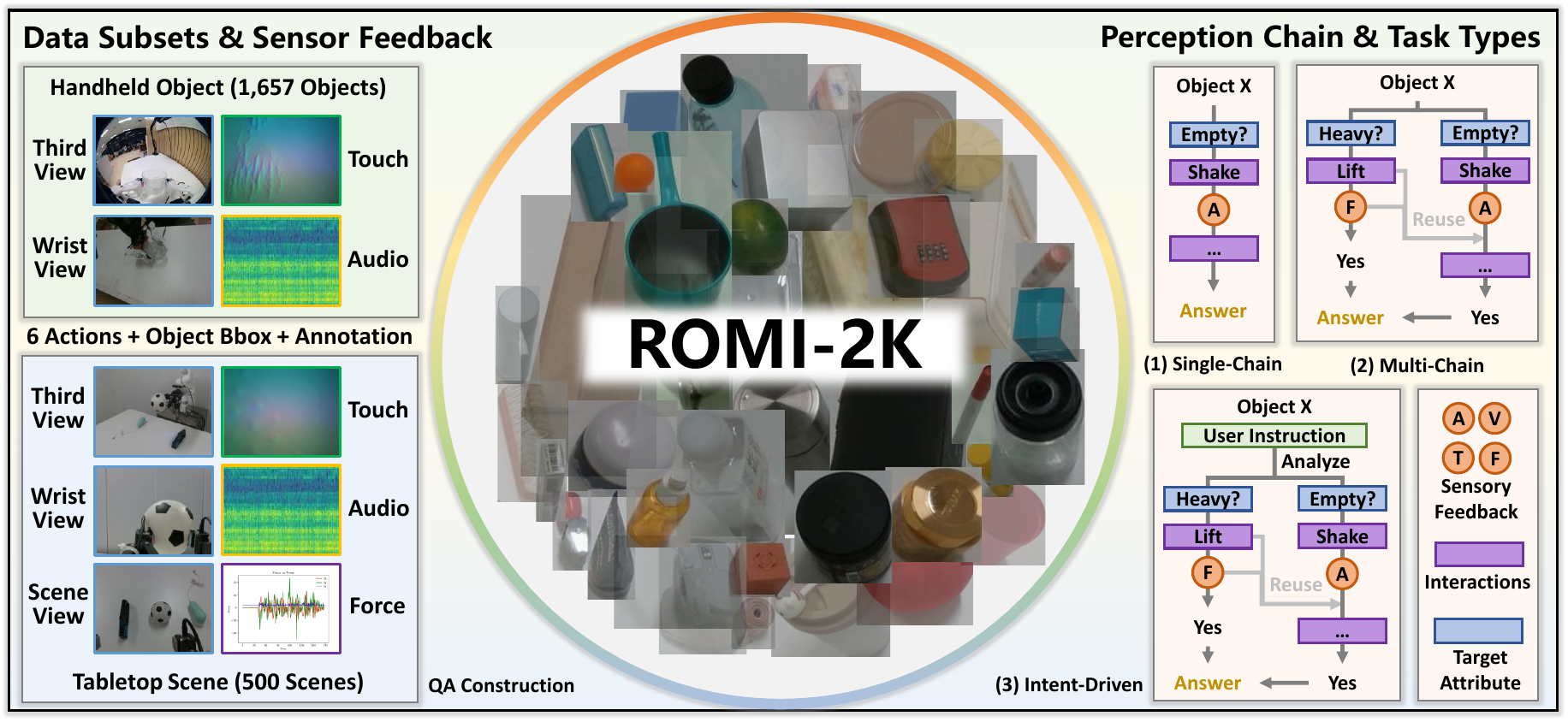}
    \caption{\textbf{Overview of ROMI-2K and the formulation of multi-sensory active perception tasks}. ROMI-2K combines two complementary real-world data subsets: a large-scale handheld object collection for diverse single-object interactions and a robotic-arm tabletop collection for multi-object scenes, with synchronized visual, audio, tactile, and force feedback. We organize the active perception tasks into perception chains and construct three task types: (1) Single-Chain tasks for an individual attribute, (2) Multi-Chain tasks for multiple attributes requiring interaction planning, and (3) Intent-Driven tasks for inferring and perceiving target attributes from implicit user instructions. Multiple perception chains may also involve reusing feedback across chains.}
    \label{fig:dataset}
\end{figure}

Existing multi-sensory interaction datasets face a fundamental trade-off between \textit{object diversity} and \textit{interaction diversity}. Interaction-centric datasets collect diverse trajectories over a limited set of objects (\textit{e.g.}, 381 objects in Hoi!~\citep{engelbracht2026hoi}). In contrast, object-centric datasets cover numerous objects with simple interactions like hitting~\citep{clarke2025x}.
Consequently, existing resources provide limited coverage of how diverse objects respond to different physical interactions and their multi-sensory feedback. This limits their suitability for real-world active perception.

To bridge this gap, we introduce \textbf{ROMI-2K}, a large-scale real-world multi-sensory object interaction dataset covering nearly 2,000 everyday objects, including diverse combinations of inside contents. For each object, we perform six atomic interactions including \textit{lift, press, collide, shake, rotate,} and \textit{squeeze}, while synchronously recording visual, audio, tactile and force observations. 
These interactions provide complementary physical cues and sensory feedback, enabling diverse object-interaction-feedback combinations for active perception.
To scale data collection while preserving realistic configurations, we adopt two complementary approaches: handheld collection for broad object-level coverage and robotic-arm collection for structured scene-level data, as shown in Fig.~\ref{fig:dataset}.

\textbf{Handheld Object Collection.} To efficiently scale object-level interaction data, we modify the UMI device Pika by integrating a microphone and two GelSight Mini~\citep{li2014GELSIGHTMNIN} sensors. The setup captures wrist-mounted and third-person visual observations with synchronized audio and tactile feedback across six atomic interactions. For each object, we perform interactions from 3 distinct grasping locations to capture diverse contact configurations and interaction outcomes. We also vary the contents of objects to collect diverse object-content combinations, particularly capturing audio feedback from shaking and rotating. This subset contains 1,657 object-content combinations, providing broad object-level coverage of physical interactions and their multi-sensory responses.

\textbf{Tabletop Scene Collection.} To study active perception in realistic multi-object environments, we collect tabletop scenes using a robotic arm. We randomly place 1--5 objects on a tabletop, with each object either empty or containing one item. We programmatically control the robot to sequentially grasp and perform the six interactions on each object, while recording synchronized visual (RGB and initial depth), audio, tactile, and force feedback. Most objects are reused from the handheld collection, forming 400 training scenes, while 100 test scenes include 56 additional unseen objects. This subset provides structured scene-level interaction data for learning and evaluating models that actively select interactions and acquire task-relevant sensory evidence in real-world environments.

We use Gemini 3.5 Flash~\citep{google2026gemini35} to annotate objects in both subsets with object identities, physical attributes, contents, and interaction-specific properties, including material, hardness, roughness, and texture. We manually verify and correct the annotations, and use them to construct scene-level active perception tasks by associating target attributes with the interactions and sensory feedback required to infer them.
For each target attribute, the required sequence of physical interactions and corresponding multi-sensory feedback forms a \textit{perception chain}, as shown in Fig.~\ref{fig:dataset}. 
Each interaction provides new sensory evidence that can influence subsequent interactions, forming a chain of evidence acquisition and reasoning.
We consider three types of active perception QA tasks with different perception structures: \textit{single-chain} tasks for single-attribute perception, \textit{multi-chain} tasks for multiple attributes requiring complementary evidence, and \textit{intent-driven} tasks for inferring and perceiving target attributes from implicit user intent. 
In multi-chain and intent-driven tasks, different perception chains may share interactions or perception modalities, requiring the model to reuse acquired evidence and avoid redundant interactions. This naturally introduces additional planning challenges in selecting informative interactions and modalities across multiple perception chains. The task difficulty is jointly determined by the number of objects, perception chain length, and task type.
We refer to the test set of the tabletop subset as \textbf{ROMA Bench}, providing a standardized offline evaluation of active perception on scene-level tasks without direct access to the robotic arm.
More details are shown in Appendix~\ref{app:handheld} and~\ref{app:Tabletop}.

\begin{figure}[t]
    \centering
    \includegraphics[width=\linewidth]{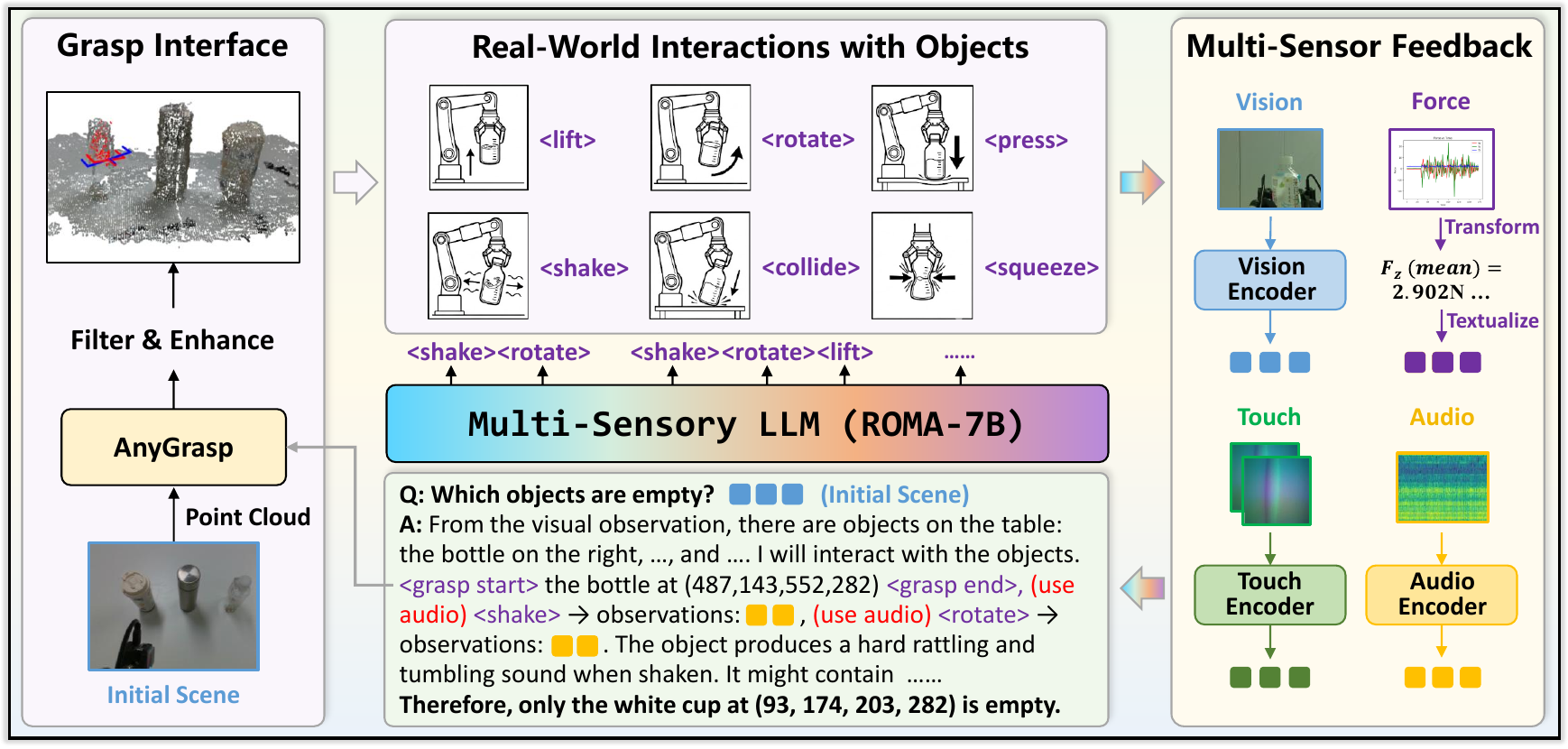}
    \caption{\textbf{Overview of the ROMA system for real-world object-centric multi-sensory active perception}. Given an initial scene and a task instruction, the Multi-Sensory LLM identifies the objects, determines the missing evidence, and selects the target object, interaction, and modality needed to acquire it. The physical interface, consisting of a grasp interface and 6 pre-defined interactions, translates these decisions into executable robot actions and collects the visual, audio, tactile, and force feedback. The acquired feedback is returned to the LLM for further reasoning, forming an iterative \textit{reasoning-interaction-feedback} loop until sufficient evidence is obtained to answer.}
    \label{fig:model}
\end{figure}

\section{REAL-WORLD Object Multi-Sensory Active Perception System}
Building on the rich object and scene interaction data provided by ROMI-2K, we introduce \textbf{ROMA}, an LLM-Based System for Real-World Object-Centric Multi-Sensory Active Perception, as shown in Fig.~\ref{fig:model}. We describe its physical interface for executing interactions and collecting multi-sensory feedback (Sec.~\ref{sec:grasp}) and its multi-sensory LLM for active perception (Sec.~\ref{sec:model}) in this section.

\subsection{Physical Interface and Multi-Sensory Feedback}
\label{sec:grasp}
Unlike passive or visual active perception, multi-sensory active perception requires physical interactions to acquire contact-dependent modalities such as touch and force. Therefore, a robust grasping and interaction interface is essential for reliably acquiring multi-sensory feedback.
Considering both cost and grasping performance, we adopt the training-free AnyGrasp~\citep{fang2023anygrasp} as our grasping policy. 
Given the 3D point cloud of an object, AnyGrasp generates a set of grasp poses ranked by confidence. However, active perception imposes stricter requirements than conventional pick-and-place: a grasp must provide stable contact for tactile sensing and remain secure throughout physical interactions. Moreover, incomplete or corrupted depth observations can make the point-cloud-based grasping unreliable for transparent and reflective objects.
To address these challenges, we first introduce a lightweight symmetry-based point cloud completion method to recover missing geometric information. We then filter and refine the generated grasp poses based on object geometry and robot constraints. Specifically, we filter infeasible candidates using object height, workspace reachability, and grasp orientation, and then refine the grasp geometry and orientation and compensate for the non-parallel fingertip motion of our gripper. These steps produce stable grasp poses suitable for multi-sensory physical interactions. The detailed grasp interface is presented in Appendix~\ref{app:graspmodule}.

With reliable grasp poses established, ROMA collects multi-sensory feedback during the 6 pre-defined interactions in Sec.~\ref{sec:dataset}. For each interaction, the robot synchronously records visual, audio, tactile, and force signals. We encode one wrist-camera image, one cropped audio segment, and two optical tactile images as optional multi-sensory feedback. Since the force sensor is mounted at the wrist, its raw data is expressed in a sensor frame determined by the gripper pose, making it difficult for the model to understand. Therefore, we transform the raw force data into the robot base frame and compensate for the empty-gripper force:
\begin{equation}
 F_{trans} = R_{\mathrm{B}\leftarrow \mathrm{G}}R_{\mathrm{G}\leftarrow \mathrm{S}}(F_{raw} - F^0_{raw}),
\end{equation}
where $F_{raw}$ denotes the raw force data, $F^0_{raw}$ denotes the force with the empty gripper, $R_{\mathrm{G}\leftarrow \mathrm{S}}$ and $R_{\mathrm{B}\leftarrow \mathrm{G}}$ represent the rotation matrices from the sensor frame to the gripper frame and from the gripper frame to the robot base frame, respectively. The resulting $F_{trans}$ represents the force of the grasped object in the base coordinate frame, providing a representation that can be directly understood by the model. We provide the averaged gravity force during each interaction as textual evidence, while leaving forces along other directions for future exploration.

\begin{figure}[t]
    \centering
    \includegraphics[width=0.8\linewidth]{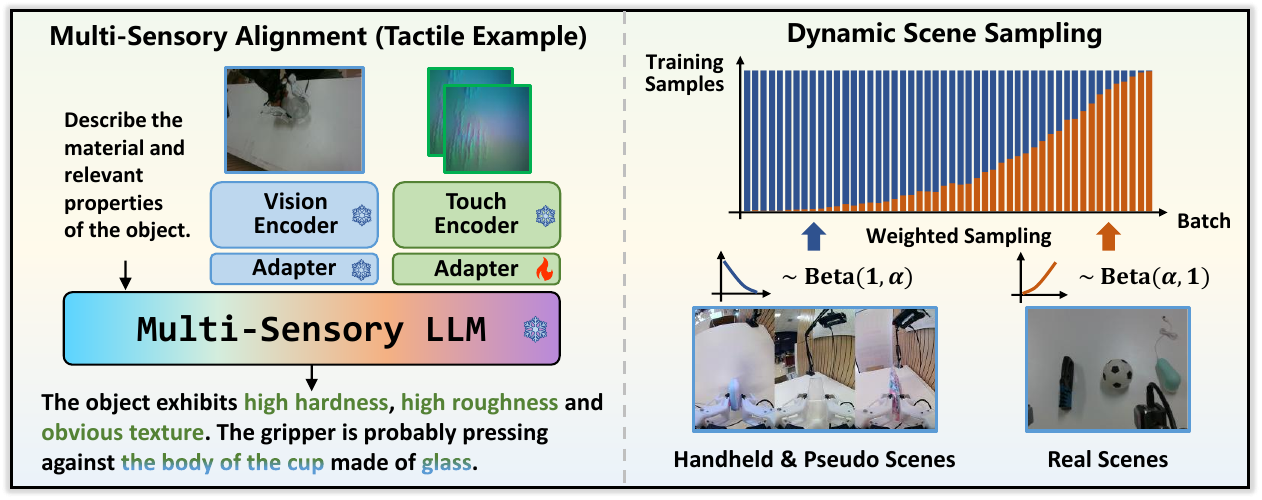}
    \caption{\textbf{Multi-Sensory Alignment and dynamic scene sampling for training the Multi-Sensory LLM}. We illustrate the alignment process using tactile-visual-text alignment as an example. We align tactile representations with visual and textual semantics, enabling the model to better leverage complementary cues across modalities. For dynamic scene sampling, we use Beta-distribution-based sampling to smoothly shift the training from interaction-level supervision toward real-world scene training.}
    \label{fig:train}
\end{figure}
\subsection{Multi-Sensory Active Perception LLM}
\label{sec:model}
The core of ROMA is to enable an LLM to identify missing evidence from the current observations and task instructions, and determine how to acquire it through physical interaction. Hence, we build our Multi-Sensory LLM ROMA-7B upon Qwen 2.5-Omni~\citep{xu2025qwen25omnitechnicalreport}, leveraging its unified vision, audio, and language understanding capabilities as the foundation for multi-sensory perception. We further extend the model with action and modality tokens, and then adopt a two-stage training strategy. Specifically, we first align the newly introduced sensory modalities, and then develop the model's ability to actively select interactions and modalities for acquiring and reasoning over the multi-sensory feedback through supervised fine-tuning (SFT), as shown in Fig.~\ref{fig:train}.

\textbf{Action and Modality Tokens.} We first introduce special tokens to explicitly represent interactions and modalities. The action tokens correspond to the 6 pre-defined interactions: \texttt{<lift>}, \texttt{<press>}, \texttt{<collide>}, \texttt{<shake>}, \texttt{<rotate>}, and \texttt{<squeeze>}. For grasping, we use \texttt{<grasp\_start>} and \texttt{<grasp\_end>} to delimit the object and its bounding box, providing the physical interface with an explicit target. We also introduce modality positional tokens for the touch and force modalities to indicate their positions and identities. Inspired by tool tokens~\citep{li2025re}, we initialize each special token by averaging the input and output embeddings of the corresponding words (\textit{e.g.}, ``shake'' $\hookrightarrow$ \texttt{<shake>}, ``grasp start'' $\hookrightarrow$ \texttt{<grasp\_start>}) and adding Gaussian noise. This semantic initialization facilitates integration of the newly introduced tokens into the pre-trained representation space.

\textbf{Stage 1: Multi-Sensory Alignment.} Before active perception training, we align the newly introduced tactile modality and the audio modality, which has limited exposure to large-scale object-centric data, with the visual-textual representation space. Specifically, we jointly perform audio-text and audio-visual-text alignment for audio, and tactile-text and tactile-visual-text alignment for touch, using instruction tuning~\citep{liu2023visual}. Such multi-modal alignment grounds non-visual sensory cues in object-level semantics, as audio and tactile signals can be ambiguous in isolation. For example, collision sounds from a plastic box and a plastic cup may be acoustically similar, whereas their visual appearances provide distinct object-level cues. This stage establishes a unified semantic space for heterogeneous sensory inputs, providing a foundation for subsequent multi-sensory reasoning.

\textbf{Stage 2: Active Perception SFT.} We then perform supervised fine-tuning to train the model to actively acquire and reason over multi-sensory evidence. Each training sample contains a task instruction, initial scene observation, and corresponding perception chains. The model first analyzes the initial scene and determines whether the available evidence is sufficient. When additional evidence is needed, it generates the target object, interaction, and sensory modalities for the next perception step. After receiving the multi-sensory feedback, it updates its reasoning and continues until sufficient evidence is obtained.
We construct the SFT data from three sources: handheld object interactions, pseudo-scenes, and real-world scenes. Handheld data provide diverse interaction-level supervision for learning action-feedback correspondences. Pseudo-scenes combine multiple objects and interaction observations to introduce object selection and perception-chain construction in more complex settings. Real-world scene data further expose the model to realistic active perception tasks and the force modality. We use a Beta-distribution-based sampling strategy to progressively shift the training distribution from interaction-level data to real-world scenes. Specifically, handheld and pseudo-scene samples are assigned positions sampled from $\mathrm{Beta}(1,\alpha)$, while real-world scene samples are assigned positions from $\mathrm{Beta}(\alpha,1)$, where $\alpha>1$. Sorting samples by these positions biases interaction-level data toward early training and real-world scenes toward later training, enabling a smooth transition from basic interaction supervision to realistic scene-level active perception.

\textbf{Inference.} During inference, we treat the action tokens as stop tokens to obtain the model's interaction decision at each step. For offline benchmark evaluation, when the model selects a grasp action, we compare its predicted bounding box with the ground-truth boxes of all objects in the scene and select the object with the highest overlap as the interaction target. We then provide the corresponding multi-sensory feedback according to the predicted interaction token and sensory modalities, and iteratively continue the perception process until sufficient evidence is acquired to answer.

Overall, ROMA combines a Multi-Sensory LLM that identifies missing evidence, selects how to acquire it, and reasons over the resulting feedback with a physical interface that executes the selected interactions and collects multi-sensory feedback. Together, they form a reasoning-interaction-feedback loop that enables ROMA to continuously acquire task-relevant sensory evidence in the real world until the task can be resolved. More system details about the physical interface and our Multi-Sensory LLM are provided in Appendix~\ref{app:Tabletop} and \ref{app:Implementation}.

\section{Experiments}
In this section, we first introduce ROMA Bench and the evaluation protocol for multi-sensory active perception (Sec.~\ref{sec:benchmark proto}). 
We then evaluate ROMA against existing MLLMs on both the offline benchmark and real-world tabletop scenes (Sec.~\ref{sec:result}), and analyze how different methods allocate interaction effort for evidence acquisition (Sec.~\ref{sec:how Interact}). More details are shown in Appendix~\ref{app:ablation}.

\subsection{ROMA Bench and Evaluation Protocol}
\label{sec:benchmark proto}
We construct ROMA Bench from the test split of the tabletop subset in ROMI-2K, comprising 2,100 scene-level active perception tasks. Each task provides an initial scene observation and task instruction, requiring the model to identify relevant objects, determine missing information, and acquire sensory evidence through interactions. The benchmark covers six object attributes: hardness (Ha.), roughness (Ro.), texture (Te.), inside contents (In.), material (Ma.), and weight (We.), testing the model's ability to select and leverage different physical interactions and sensory modalities. We categorize the tasks into Single-Chain, Multi-Chain, and Intent-Driven settings, as described in Sec.~\ref{sec:dataset}, and formulate them as multiple-choice, true-or-false, and ranking questions. These tasks evaluate active perception across different levels of evidence acquisition and reasoning complexity. We provide all baselines with a standardized system prompt describing the task setting, available interactions, and sensory modalities. During inference, we provide sensory feedback based on the predicted interaction and modalities, and match each predicted grasp bounding box to the ground-truth object with the highest overlap. If all overlaps are below $70\%$, the task is considered unsuccessful. The process continues until the model terminates the interaction process and produces a final prediction, which is then compared with the ground truth using a programmatic evaluator. 
We consider a task \textit{completed} if the model terminates the interaction and reasoning process, and produces a final prediction, regardless of whether the prediction is correct. 
More statistics of ROMA Bench are shown in Appendix~\ref{app:dataset}.

\begin{table}[t]
\centering
\small
\caption{Evaluation of multi-sensory active perception on ROMA Bench across three task types. Har. (Hardness), Rou. (Roughness), Tex. (Texture), Ins. (Inside), Mat. (Material), and Wei. (Weight) denote the target object attributes in single-chain and multi-chain perception tasks. We report task success rate, where failures in object bounding box localization are also counted as task failures. \textbf{Bold} and \underline{underlined} values indicate the best and second-best results, respectively.}
\label{tab:main_results}
\tabcolsep=0.065cm
\begin{tabular}{c|ccccccc|ccccccc|c|c}
\toprule
\multirow{2}{*}{\textbf{Model}}
& \multicolumn{7}{c|}{\textbf{Single-Chain}}
& \multicolumn{7}{c|}{\textbf{Multi-Chain}}
& \multirow{2}{*}{\makecell{\textbf{Intent-}\\\textbf{Driven}}}
& \multirow{2}{*}{\textbf{Total}} \\
\cmidrule(lr){2-8}
\cmidrule(lr){9-15}
& \textbf{Har.} & \textbf{Rou.} & \textbf{Tex.} & \textbf{Ins.} & \textbf{Mat.} & \textbf{Wei.} & \textbf{All}
& \textbf{Har.} & \textbf{Rou.} & \textbf{Tex.} & \textbf{Ins.} & \textbf{Mat.} & \textbf{Wei.} & \textbf{All}
& & \\
\midrule

GPT-5.4
& 34.3 & 41.0 & 38.1 & 23.1 & 60.1 & 38.6 & 45.1
& 32.0 & 37.6 & 38.1 & 41.3 & 42.1 & 35.8 & 38.4
& 56.0 & 43.5 \\

GPT-6 Astra
& \textbf{70.7} & \textbf{88.5} & \underline{76.2} & \underline{32.1} & \textbf{75.9} & \textbf{100.0} & \textbf{76.9}
& \textbf{62.8} & \underline{72.1} & \underline{65.9} & \underline{56.3} & \underline{68.6} & \underline{75.2} & \underline{66.6}
& \textbf{79.9} & \underline{72.3} \\

Gemini 3.5 Flash
& 45.5 & 66.7 & 50.0 & 28.2 & 58.4 & 60.8 & 54.4
& 48.3 & 58.8 & 46.9 & 46.4 & 51.8 & 52.9 & 49.9
& 59.4 & 53.0 \\

Qwen 2.5-Omni
& 19.2 & 12.8 & 9.5 & 5.1 & 37.8 & 5.9 & 21.1
& 20.1 & 16.4 & 18.8 & 23.3 & 28.8 & 15.0 & 21.2
& 26.3 & 22.0 \\

Qwen 3-Omni
& 4.0 & 14.1 & 19.0 & 6.4 & 53.3 & 0.0 & 24.7
& 12.3 & 10.3 & 16.6 & 16.8 & 22.0 & 0.7 & 13.3
& 28.8 & 19.7 \\

\textbf{ROMA-7B}
& \underline{48.5} & \underline{67.9} & \textbf{81.0} & \textbf{59.0} & \underline{70.8} & \underline{91.5} & \underline{71.1}
& \underline{62.5} & \textbf{75.8} & \textbf{77.4} & \textbf{75.1} & \textbf{70.7} & \textbf{80.9} & \textbf{74.1}
& \underline{73.1} & \textbf{72.9} \\

\bottomrule
\end{tabular}
\end{table}

\begin{table}[t]
\centering
\small
\caption{Evaluation of multi-sensory active perception in real-world scenes using our physical interface across single-chain, multi-chain, and intent-driven tasks. Unsuccessful grasps caused by object localization failures are also counted as task failures. \textbf{Bold} and \underline{underlined} values indicate the best and second-best results,  respectively.}
\label{tab:real_world_results}
\tabcolsep=0.35cm
\begin{tabular}{c|cccc}
\toprule
\textbf{Model}
& \textbf{Single-Chain}
& \textbf{Multi-Chain}
& \textbf{Intent-Driven}
& \textbf{Total} \\
\midrule

GPT-5.4
& 42.6 & 29.4 & 55.6 & 40.2 \\

GPT-6 Astra
& \textbf{64.8} & \underline{60.8} & \textbf{66.7} & \textbf{63.6} \\

Gemini 3.5 Flash
& 50.0 & 29.4 & 48.1 & 41.7 \\

Qwen 2.5-Omni
& 20.4 & 29.4 & 18.5 & 23.5 \\

Qwen 3-Omni
& 33.3 & 15.7 & 18.5 & 23.5 \\

\textbf{ROMA-7B}
& \underline{57.4} &\textbf{ 64.7} & \underline{63.0} & \underline{61.4} \\

\bottomrule
\end{tabular}
\end{table}

\begin{figure}[t]
    \centering
    \includegraphics[width=\linewidth]{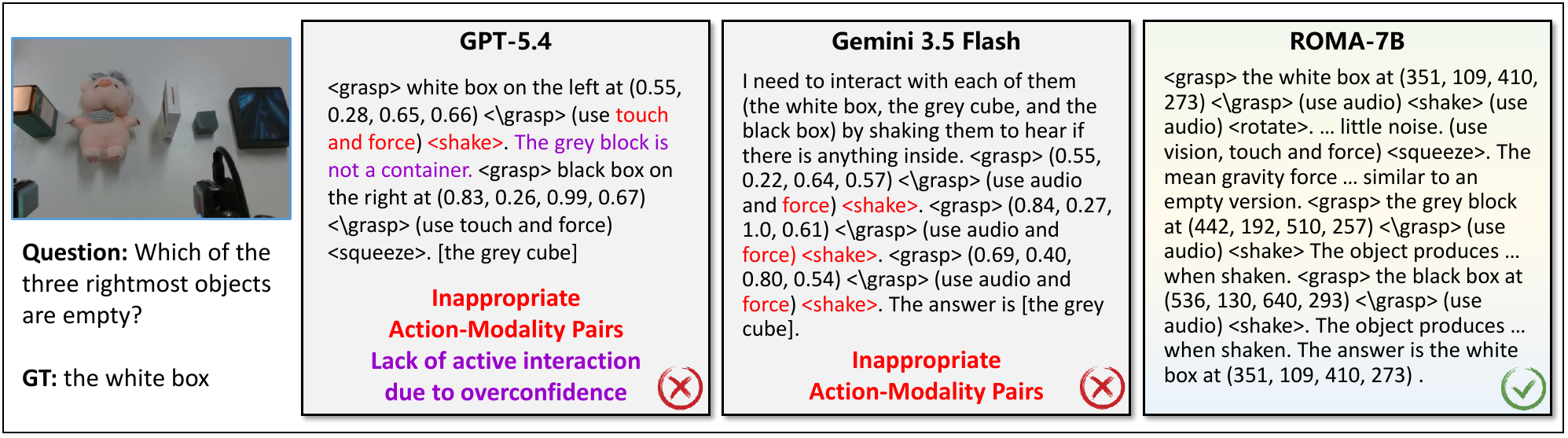}
    \caption{A case study of MLLM baselines and ROMA-7B on multi-sensory active perception tasks in ROMA Bench. GPT-5.4 tends to terminate with fewer active interactions, occasionally skipping critical objects and sensory evidence. Both GPT-5.4 and Gemini 3.5 Flash also produce inappropriate action-modality combinations, such as attempting to observe gravity force during vigorous shaking. In contrast, ROMA-7B more often continues interaction when the available evidence is insufficient and selects complementary actions and modalities to acquire task-relevant information.}
    \label{fig:case}
\end{figure}

\subsection{Can ROMA Effectively Perform Active Perception?}
\label{sec:result}
We first evaluate whether ROMA can actively identify missing evidence, select appropriate interactions and sensory modalities, and reason over the resulting feedback on ROMA Bench. As shown in Tab.~\ref{tab:main_results}, ROMA-7B substantially outperforms existing MLLM baselines and remains highly competitive with the much larger and more recent GPT-6 Astra, achieving 72.9\% overall accuracy compared with 72.3\% for GPT-6 Astra. This is particularly notable given the substantially smaller model size of ROMA-7B. GPT-6 Astra exhibits strong visual perception and reasoning capabilities, achieving higher performance than ROMA-7B on Single-Chain and Intent-Driven tasks. It also rarely suffers from object localization errors, suggesting strong visual grounding in the active perception setting. Moreover, GPT-6 achieves strong performance on several attributes even when tactile observations are provided through a visual interface, in some cases exceeding ROMA-7B with its dedicated AnyTouch~2-based tactile perception. Nevertheless, ROMA-7B maintains a slight overall advantage, demonstrating the effectiveness of ROMI-2K in providing interaction-grounded multi-sensory supervision for learning to actively acquire and integrate complementary physical evidence.

We further conduct real-world embodied evaluation by randomly selecting eight scenes from ROMA Bench due to the high cost of physical execution, and recreating them in the real world. This results in 132 tasks, providing a sufficiently large scale for real-world embodied evaluation. Each task is executed on a real robotic arm, and answer options are removed to require free-form responses.
The correctness of the responses is manually verified. All models use the same physical interface from ROMA for physical execution and sensory feedback collection. Real-world execution introduces additional challenges, including object localization for actual grasping and sensory noise from execution errors. As shown in Tab.~\ref{tab:real_world_results}, GPT-6 Astra achieves the highest overall real-world performance at 63.6\%, closely followed by ROMA-7B at 61.4\%. Notably, ROMA-7B remains the best-performing model on Multi-Chain tasks, consistent with its advantage in integrating evidence across multiple interaction steps. Although ROMA-7B slightly trails behind GPT-6 Astra, it outperforms the frontier GPT-5.4 and Gemini 3.5 Flash, despite being built on the 7B Qwen2.5-Omni backbone with substantially weaker performance on active perception. This demonstrates that ROMA's multi-sensory active perception capability remains effective under real-world physical execution.

We also identify two recurring issues in baseline failures: object localization and interaction-specific sensory reasoning. Although existing MLLMs exhibit strong visual and semantic understanding, GPT-5.4 and Gemini 3.5 Flash often struggle to localize target objects in the more complex active perception setting, where localization is an important component of active perception and is coupled with interaction selection and multi-sensory evidence acquisition.
They also show limited understanding of interaction-dependent sensory semantics, which remains challenging even for GPT-6 Astra. For example, models may incorrectly interpret the force measured during squeezing as grasping force or attempt to acquire gravity force feedback during vigorous shaking, leading to inappropriate action-modality choices and subsequent reasoning errors.
In addition, inside-related tasks remain particularly challenging. They primarily rely on audio cues but often require complementary evidence from other sensory modalities. Solving these tasks therefore requires integrating multiple sensory observations with multi-modal knowledge about the object and its contents.
As illustrated by the representative case in Fig.~\ref{fig:case}, these examples highlight the limitations of existing MLLMs and the benefits of our work: ROMI-2K provides interaction-grounded multi-sensory supervision that existing datasets lack, while ROMA enables active selection of interactions and sensory modalities to acquire task-relevant evidence beyond passive multi-sensory reasoning.
More detailed ablation studies are shown in Appendix~\ref{app:ablation}.

\begin{table}[t]
\centering
\small
\caption{
Comparison of grasp and interaction counts on completed active perception tasks. For ROMA Bench, oracle bounding boxes are provided to isolate interaction behavior from object localization errors. \textit{Per Scene} denotes the average number of grasps or further interactions per scene, while \textit{Per Object} denotes the average number of further interactions per grasped object. \textit{Exhaustive} grasps every object and executes all six interactions on each object without active selection.
}
\label{tab:interaction_counts}
\tabcolsep=0.17cm
\begin{tabular}{c|cccc|ccc}
\toprule
\multirow{2}{*}{\textbf{Method}}
& \multicolumn{4}{c|}{\textbf{ROMA Bench (w/ Oracle Bounding Box)}}
& \multicolumn{3}{c}{\textbf{Real World}} \\
\cmidrule(lr){2-5}
\cmidrule(lr){6-8}
& \textbf{Grasp}
& \multicolumn{2}{c}{\textbf{Further Interaction}}
& \multirow{2}{*}{\makecell{\textbf{Total}\\\textbf{Acc.}}}
& \textbf{Grasp}
& \multicolumn{2}{c}{\textbf{Further Interaction}} \\
& \textbf{Per Scene}
& \textbf{Per Scene}
& \textbf{Per Object}
&
& \textbf{Per Scene}
& \textbf{Per Scene}
& \textbf{Per Object} \\

\midrule
\noalign{\vskip -2.1pt}
\rowcolor{gray!15}
Exhaustive
& 3.68 & 22.09 & 6.00 & --
& 3.21 & 19.27 & 6.00\\
\noalign{\vskip 1pt}

GPT-5.4
& 1.10 & 1.20 & 1.09 & 63.3
& 0.63 & 0.72 & 1.14 \\

GPT-6 Astra
& 2.29 & 2.61 & 1.14 & 73.5
& 2.08 & 2.24 & 1.08 \\

Gemini 3.5 Flash
& 2.23 & 2.77 & 1.24 & 68.8
& 2.24 & 2.92 & 1.30 \\

\textbf{ROMA-7B}
& 2.75 & 5.40 & 1.97 & \textbf{76.0}
& 2.57 & 4.97 & 1.94 \\

\bottomrule
\end{tabular}
\end{table}

\subsection{How Does ROMA Interact to Acquire Evidence?}
\label{sec:how Interact}
Accuracy alone does not fully characterize active perception. We therefore analyze the grasp and interaction behavior of different models on completed tasks in both ROMA Bench and real-world scenes. 
For ROMA Bench, we provide oracle bounding boxes to isolate interactions from object localization errors.
We additionally report the grasp and interaction counts under an \textit{exhaustive} strategy as an upper-bound reference for exploration cost. We omit its accuracy because exhaustive exploration produces a multi-sensory context that is too long to be effectively handled within our training setup (4 modalities $\times$ 6 actions).
As shown in Table~\ref{tab:interaction_counts}, ROMA-7B achieves the highest accuracy while using substantially fewer grasps and interactions than exhaustive exploration in both settings. This indicates that ROMA selectively acquires additional evidence rather than exhaustively executing all available interactions. Compared with GPT-5.4 and Gemini 3.5 Flash, ROMA-7B performs more interactions on completed tasks and achieves substantially higher accuracy, suggesting that selectively extending the perception chain can provide useful additional evidence when the available information is insufficient. As illustrated by the failure case in Fig.~\ref{fig:case}, the lower interaction counts of GPT-5.4 can result from premature termination before sufficient evidence is acquired. In contrast, ROMA-7B continues the perception chain when the available evidence is insufficient, using additional interactions to acquire complementary sensory evidence before reaching a conclusion.

The comparison with GPT-6 Astra further highlights this behavior. GPT-6 Astra increases its average number of interactions from 1.20 to 2.61 per scene compared with GPT-5.4, accompanied by a substantial improvement in accuracy from 63.3\% to 73.5\%. This suggests that GPT-6 Astra is less likely to stop after limited interactions and instead actively acquires more evidence before making a decision. 
Nevertheless, ROMA-7B further extends the perception chain, performing 5.40 interactions per scene and achieving a higher accuracy. This suggests that ROMA can more effectively exploit complementary physical interactions to acquire task-relevant sensory evidence when the current observations are insufficient. This ability to actively extend the perception chain contributes to ROMA-7B's stronger performance than many frontier MLLMs and its comparable performance with GPT-6 Astra. More interaction analysis is shown in Appendix~\ref{app:ablation}.


\section{Conclusion}

In this work, we advance multi-sensory perception from passive sensory integration toward active evidence acquisition. \textbf{ROMI-2K} provides diverse real-world object interactions with visual, audio, tactile, and force feedback, enabling models to learn how physical interactions reveal object properties. Building on these data, \textbf{ROMA} actively selects target objects, interactions, and sensory modalities based on the information required by the task, and adapts its perception process based on the resulting feedback. We further establish \textbf{ROMA Bench} as a unified evaluation framework spanning different levels of evidence acquisition and reasoning complexity. Together, ROMI-2K, ROMA, and ROMA Bench establish a foundation for object-centric multi-sensory active perception, moving beyond interpreting available observations toward actively acquiring the evidence needed to understand the physical world. We hope this work inspires future embodied agents that tightly integrate perception, interaction, and reasoning for more active multi-modal intelligence. The detailed discussion of limitations and future work is provided in Appendix~\ref{app:limit}.

\section{Acknowledgements}
This work was supported in part by the Beijing Major Science and Technology Project under Contract no. Z251100008125018. This work was also supported by Beijing Academy of Artificial Intelligence (BAAI). We would also like to thank Siyu Mei and Mingxin Wang for their help with data collection and annotation.

\clearpage

\bibliographystyle{IEEEtran}
\bibliography{paper}

\clearpage

\beginappendix

\section{Appendix Introduction}
This appendix provides additional details on the design, implementation, data collection, and evaluation of ROMA. We first discuss the limitations of the current system and potential directions for future research in~\ref{app:limit}. We then describe the two complementary data collection platforms used to construct ROMI-2K, including the handheld object interaction platform and the tabletop robotic system, together with their hardware configurations, data collection procedures, annotations, and QA construction in~\ref{app:handheld} and~\ref{app:Tabletop}. We further provide detailed statistics of the ROMI-2K dataset in~\ref{app:dataset}, followed by the training configuration, data composition, and evaluation settings of ROMA in~\ref{app:Implementation}. Finally, we provide additional analyses of interaction behavior and model components, including interaction patterns, the impact of object localization errors, missing-modality settings, and component-level ablations in~\ref{app:ablation}. We also provide a demonstration video in the supplementary material.

\section{Limitations and Future Work}
\label{app:limit}
Establishing a real-world system that can actively acquire and reason over visual, audio, tactile, and force information already requires substantial effort in data collection, multi-sensory alignment, robotic interaction, grasp planning, and model training. ROMA takes an important step toward this goal by establishing a unified framework for object-centric multi-sensory active perception, together with the ROMI-2K dataset and a physical interaction system that supports diverse real-world object exploration. At the same time, the current system provides a foundation rather than a complete solution to the multi-sensory embodied agent. Its interaction space, reasoning horizon, memory, embodiment, and use of force feedback can all be further extended. We believe that the framework and data established in this work provide a strong starting point for the following directions:
\begin{itemize}
    \item \textbf{Integration with Manipulation Policies.} ROMA currently uses physical interaction primarily as a means of acquiring perceptual evidence. The resulting physical understanding can naturally serve as a foundation for downstream manipulation. For example, the material, hardness, roughness, weight, and contents inferred through active perception could guide grasp selection, manipulation strategy, and action adaptation. Building upon ROMA's existing interaction-perception-reasoning loop, a promising direction is to integrate active perception with manipulation policies into a unified interaction-perception-reasoning-action framework. Such a framework could allow the robot to actively acquire information when necessary, and then immediately use the acquired knowledge to perform more reliable and physically appropriate manipulation.
    
    \item \textbf{Integration of Interaction Skills.} ROMA currently organizes physical exploration around 6 atomic interactions: lift, press, collide, shake, rotate, and squeeze. These primitives provide a simple and general interaction space and establish a common interface between the LLM and the robotic system. A natural extension is to build upon these primitives with richer learned interaction skills that can be composed or adapted according to the object, task, and currently missing information. In addition, ROMI-2K provides a diverse collection of object-interaction trajectories that could serve as a foundation for learning such skills. Beyond learned skills, future systems could also allow users or external programs to define new interaction skills through natural language or programmatic specifications. These skills could then be introduced dynamically according to the task or environment without retraining the entire model. This would extend ROMA from selecting among a fixed set of atomic actions toward composing and executing increasingly diverse information-seeking behaviors.
    
    \item \textbf{Multi-turn Dialogue and Complex Reasoning.} The current ROMA benchmark primarily evaluates active perception under individual task instructions. In practical applications, however, users may provide incomplete requirements, refine their questions through multiple turns, or ask follow-up questions based on previously acquired evidence. The multi-sensory object interactions and object-level annotations in ROMI-2K provide a natural basis for constructing more diverse multi-turn dialogue data. For example, multiple questions about the same object can be organized into a continuous conversation, where previously acquired sensory evidence is retained and reused to answer subsequent questions. More complex tasks could further require the model to compare multiple objects, revise its perception strategy according to the dialogue history, or integrate evidence collected across multiple perception chains. Extending ROMA in this direction would enable active perception to evolve from single-task information seeking toward longer-horizon interactive reasoning.

    \item \textbf{Multi-Sensory Information Compression and Summarization.} As the number of interactions and perception steps increases, continuously retaining raw observations from multiple modalities can lead to rapidly growing sensory context. Future work could therefore investigate compact representations that preserve the most informative physical evidence while reducing redundant multi-sensory observations. For example, visual, audio, tactile, and force observations collected across an interaction sequence could be summarized into object-level physical representations or structured sensory memories. Such representations could retain key evidence about material, hardness, roughness, weight, contents, and other physical properties while discarding redundant observations. Multi-sensory information compression could also facilitate longer-horizon reasoning and persistent object memory by providing a compact interface between previously acquired physical knowledge and subsequent perception or manipulation tasks.
    
    \item \textbf{Persistent Object Memory.} ROMA currently focuses on acquiring physical knowledge within the current scene and interaction process. A natural next step is to maintain persistent object-level memory across different episodes and environments. The rich object-level physical attributes and interaction observations collected in ROMI-2K provide a useful foundation for studying how such knowledge can be stored, retrieved, and updated. In practical settings such as homes, laboratories, and other long-term environments, a robot could remember previously observed objects, their physical properties, interaction outcomes, and task-specific knowledge, and reuse this information when encountering the same or similar objects. Persistent memory could therefore reduce redundant exploration while enabling the robot to gradually build a continuously expanding physical understanding of its environment.
    
    \item \textbf{Scaling to More Embodied Platforms.} ROMA is currently instantiated on a single-arm robotic platform with a specific multi-sensory gripper. Extending the framework to different embodiments, such as dual-arm robots and dexterous hands, could substantially expand the space of available interactions. Different embodiments may also provide different contact configurations and sensing capabilities, leading to different strategies for acquiring the same physical information. The object-centric active perception framework and interaction-perception-reasoning loop developed in ROMA provide a general basis for studying how perception strategies can adapt to these different embodiments.
    
    \item \textbf{Full Integration of Force Modality.} In the current system, force feedback is represented primarily through the mean gravity force during interactions, providing a simple and interpretable signal for estimating object weight. This design enables force information to be incorporated into the current multi-sensory reasoning framework while keeping the representation straightforward for the model. The full six-axis force and torque signals collected by our robotic system, however, contain substantially richer information that remains to be explored. Future work could build upon the existing force data and multi-sensory alignment framework to directly model force and torque trajectories throughout different interactions. This could enable ROMA to reason about contact dynamics, friction, compliance, resistance, and other physical properties, further strengthening the complementary role of force sensing alongside vision, audio, and touch.
\end{itemize}

\section{Handheld Data Collection}
\label{app:handheld}
\begin{figure}[h]
    \centering
    \includegraphics[width=0.99\linewidth]{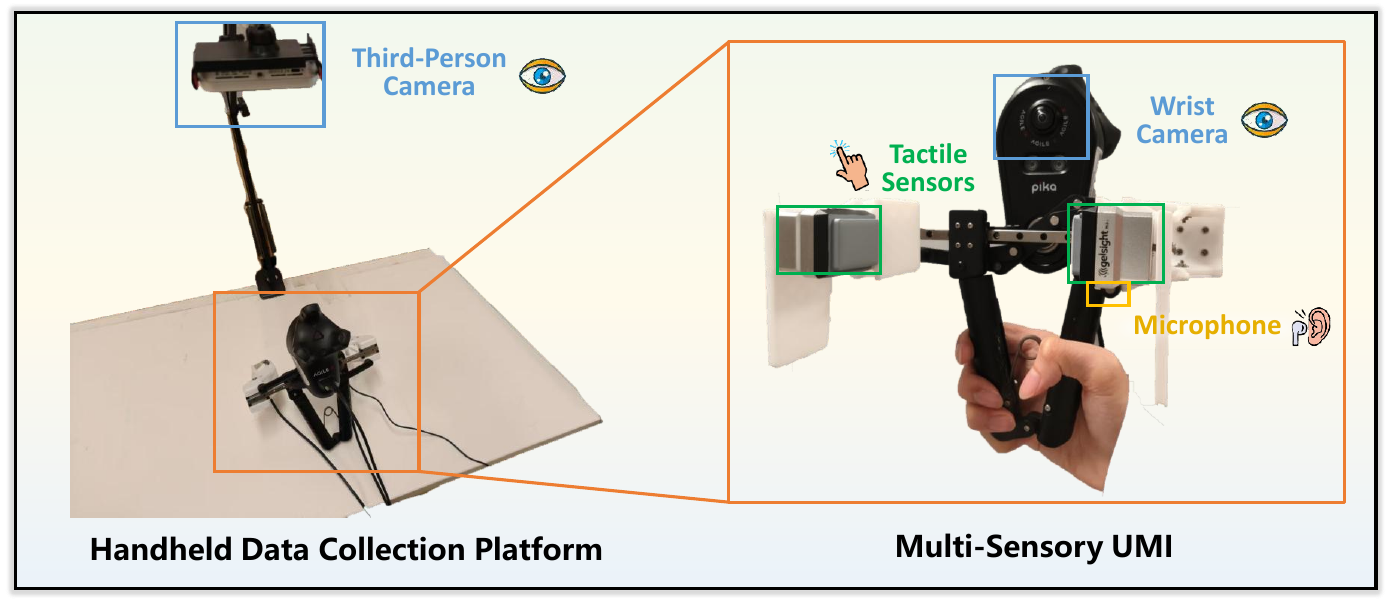}
    \caption{The illustration of the handheld object interaction data collection platform.}
    \label{fig:handplatform}
\end{figure}
Before collecting structured tabletop scenes with the robotic arm, we first build a handheld data collection platform to efficiently acquire large-scale object-interaction trajectories. The handheld platform allows us to rapidly explore a diverse set of everyday objects under different grasping and interaction configurations. These data provide broad coverage of object-level physical interactions and serve as an important component of ROMI-2K for training ROMA's active perception capabilities.

\begin{table}[t]
\centering
\small
\begin{tabular}{ccc}
\hline\noalign{\smallskip}
\textbf{Hardware} & \textbf{Model} & \textbf{Role and Frequency} \\
\noalign{\smallskip}\hline\noalign{\smallskip}
Handheld UMI & Agilex Pika Sense & Handheld Object Interaction\\
Wrist Camera & Fisheye RGB Camera $\times$ 1 & Wrist-view RGB Observation (10 fps) \\
Third-Person Camera & Intel RealSense D455f $\times$ 1 & External RGB Observation (10 fps) \\
Tactile Sensor & GelSight Mini $\times$ 2 &Optical Tactile Collection (10 fps) \\
Microphone & USB Microphone & Audio Collection \\
\hline
\end{tabular}
\caption{Hardware configuration of the handheld data collection platform.}
\label{tab:handheld-device}
\end{table}
\subsection{Hardware}
We build the handheld data collection platform based on the Agilex Pika Sense, which provides a built-in fisheye RGB camera for wrist-view visual observations. We further equip the platform with a third-person Intel RealSense D455f, two GelSight Mini tactile sensors mounted on the gripper fingers, and a USB microphone for external visual, tactile, and audio sensing, respectively. The complete hardware configuration is summarized in Tab.~\ref{tab:handheld-device}.

\subsection{Collecting Process}
For each object, we first select three different stable grasp locations that allow the object to be safely manipulated by the handheld gripper. At each grasp location, we sequentially perform the same set of six pre-defined atomic interactions: \textit{lift}, \textit{press}, \textit{collide}, \textit{shake}, \textit{rotate}, and \textit{squeeze}. 
Thus, each object is collected under three grasp configurations, with all six interactions performed once at each configuration. During each interaction, we synchronously record the wrist-view RGB image, third-person RGB image, tactile observations from both GelSight Mini sensors, and the corresponding audio signal. This provides multiple interaction trajectories for each object and captures complementary sensory responses under different grasping configurations. The six atomic interactions are defined as follows:
\begin{itemize}
\item \textbf{\textit{Lift}:} The gripper lifts the grasped object upward and holds it briefly. While this interaction provides weight-related force information on our robotic platform, the handheld Pika Sense lacks a force sensor and therefore uses lifting primarily to capture multi-sensory observations under stationary and simple-motion states.

\item \textbf{\textit{Press}:} The gripper presses the object against the table surface, mainly generating cues that reveal its hardness properties.

\item \textbf{\textit{Collide}:} The gripper repeatedly collides the grasped object with the table, capturing the resulting tactile, visual, and acoustic responses.

\item \textbf{\textit{Shake}:} The gripper repeatedly shakes the object while maintaining the grasp, producing dynamic and acoustic cues that can reveal information about its contents and internal structure.

\item \textbf{\textit{Rotate}:} The gripper rotates the grasped object to expose different surface regions and complement \textit{shake}-based acoustic cues, particularly for objects that produce weak or indistinct sounds when shaken.

\item \textbf{\textit{Squeeze}:} The gripper applies a closing motion to compress the object, providing cues about its roughness and hardness while revealing whether soft packaging contains internal objects.

\end{itemize}

To increase the diversity of the collected data, we collect objects with different shapes, materials, sizes, and physical properties. For container objects, we additionally vary their internal contents when applicable. The same container can therefore produce different visual, tactile, and acoustic responses under different contents. This results in a diverse collection of object-interaction trajectories and object-content combinations for ROMI-2K, covering 1,657 different object combinations.

\begin{figure}[h]
    \centering
    \includegraphics[width=0.9\linewidth]{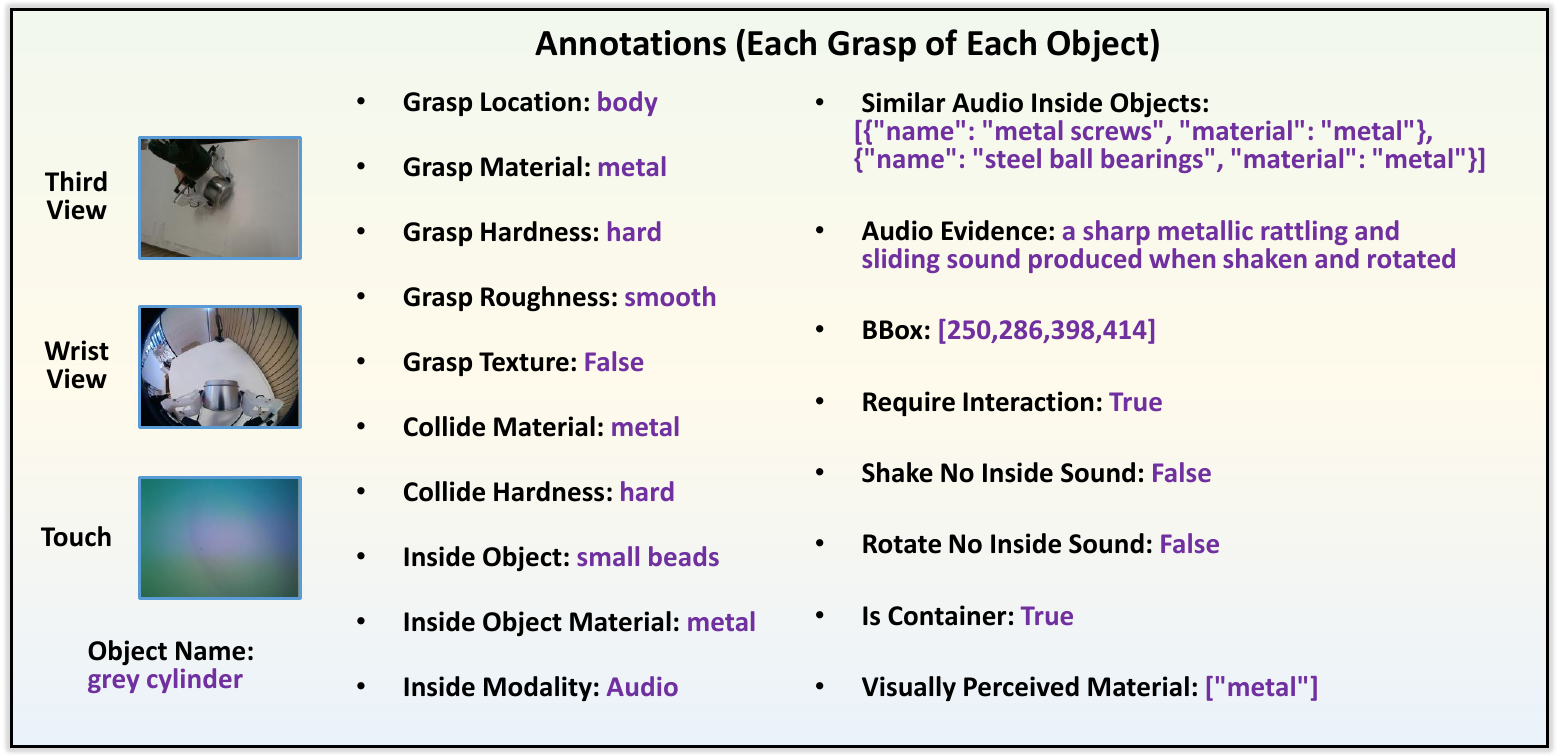}
    \caption{An example of a multi-sensory data sample and its annotations from the handheld object subset of ROMI-2K.}
    \label{fig:hand anno}
\end{figure}
\subsection{Annotation}
\label{app:hand anno}
We annotate each collected object with its name and physical attributes, including \textit{material}, \textit{hardness}, \textit{roughness}, and \textit{texture}, as well as the \textit{visually perceived material} inferred from its appearance. The latter captures what material the object appears to be made of from visual observations, which may differ from its actual material. For each of the three grasp locations, we further annotate the local \textit{material}, \textit{hardness}, \textit{roughness}, and \textit{texture} at the grasped region. For each collision location, we annotate the corresponding local \textit{material} and \textit{hardness}. These annotations characterize both the global properties of an object and the local properties exposed through different interaction locations.
For hardness and roughness, we categorize objects into three levels. Hardness is categorized as hard, medium, or soft. Hard objects exhibit little to no deformation under squeezing, such as most metal objects; medium-hard objects exhibit slight deformation, such as paper boxes; and soft objects exhibit substantial deformation, such as sponges. Roughness is similarly categorized as rough, medium, or smooth. Rough objects exhibit strong friction or resistance during touch, such as coarse ceramic objects; medium objects exhibit relatively weak friction, such as cardboard boxes; and smooth objects exhibit little perceptible friction, such as porcelain objects. Texture is annotated as a binary attribute: true indicates the presence of clearly discernible regular patterns, symbols, or markings, while false indicates their absence.
For objects that contain internal objects, we provide a more detailed set of contents-related annotations. We record the \textit{inside object} and whether the object is a typical container. We further annotate which sensory modalities can potentially provide evidence of the presence of internal contents and the physical reasons underlying such evidence when applicable. For audio, we additionally annotate whether the \textit{shake} and \textit{rotate} recordings contain audible cues from the internal object, as well as other objects that may produce acoustically similar sounds (\textit{similar audio inside object}). We also indicate whether physical interaction is necessary to determine the presence of internal contents from the available observations.
These annotations explicitly connect object properties and interaction outcomes to the sensory evidence required for active perception. They provide supervision for learning when an object can be identified from passive observations and when additional interaction is necessary to reveal otherwise inaccessible information.

We first use Gemini 3.5 Flash to generate preliminary annotations from the collected object information and multi-sensory observations. The generated annotations are then manually verified and corrected to ensure consistency with the objects and their recorded interaction trajectories. This process provides reliable object- and interaction-level labels while reducing the cost of manual annotation. An example of the annotation is shown in Fig.~\ref{fig:hand anno}.

\subsection{QA Construction}
\label{app:hand qa}
\begin{figure}[h]
    \centering
    \includegraphics[width=0.9\linewidth]{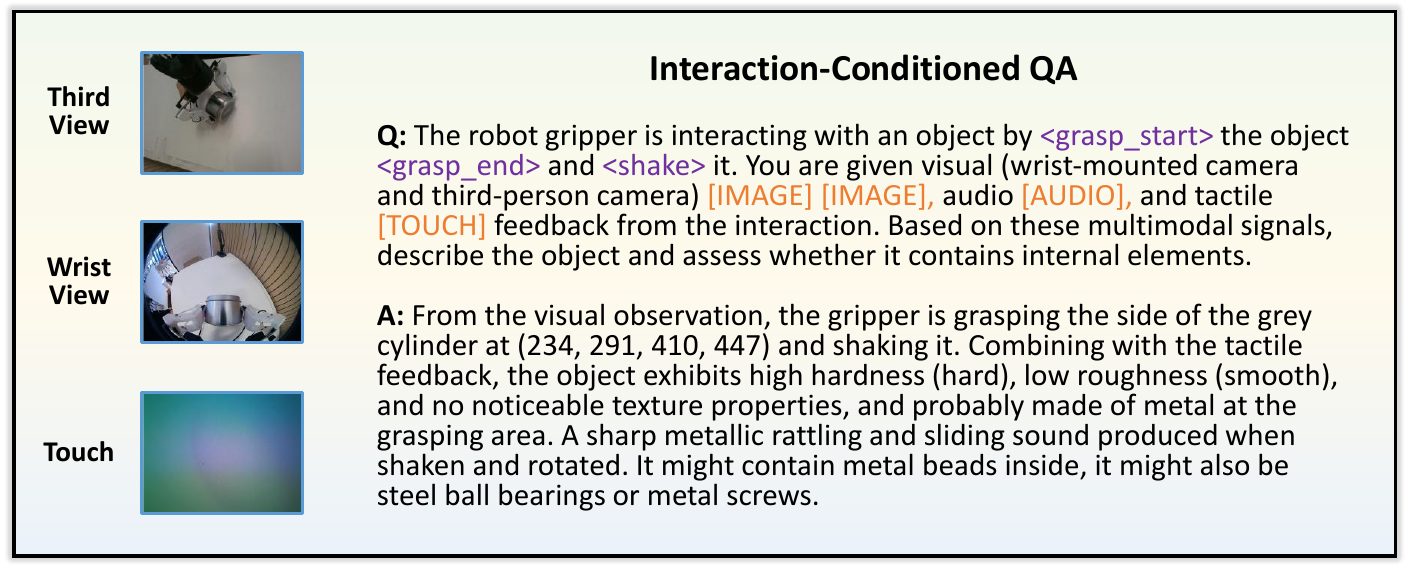}
    \caption{An example of an interaction-conditioned QA sample constructed using the handheld object subset of ROMI-2K.}
    \label{fig:hand qa}
\end{figure}
To construct training QA pairs from the handheld object data, we consider two different types of QA tasks: interaction-conditioned QA and active perception QA. We first construct interaction-conditioned QA pairs directly from the data collected for individual objects. An example is shown in Fig.~\ref{fig:hand qa}. For each object and each available interaction, we provide the model with the object observation, the ongoing interaction, and the corresponding visual, audio, and tactile observations, and ask it to answer questions about various object-related properties. These samples explicitly associate each physical interaction with its resulting multi-sensory observations and the corresponding object information, providing supervision for learning how different interactions reveal complementary physical properties.

\begin{figure}[h]
    \centering
    \includegraphics[width=0.9\linewidth]{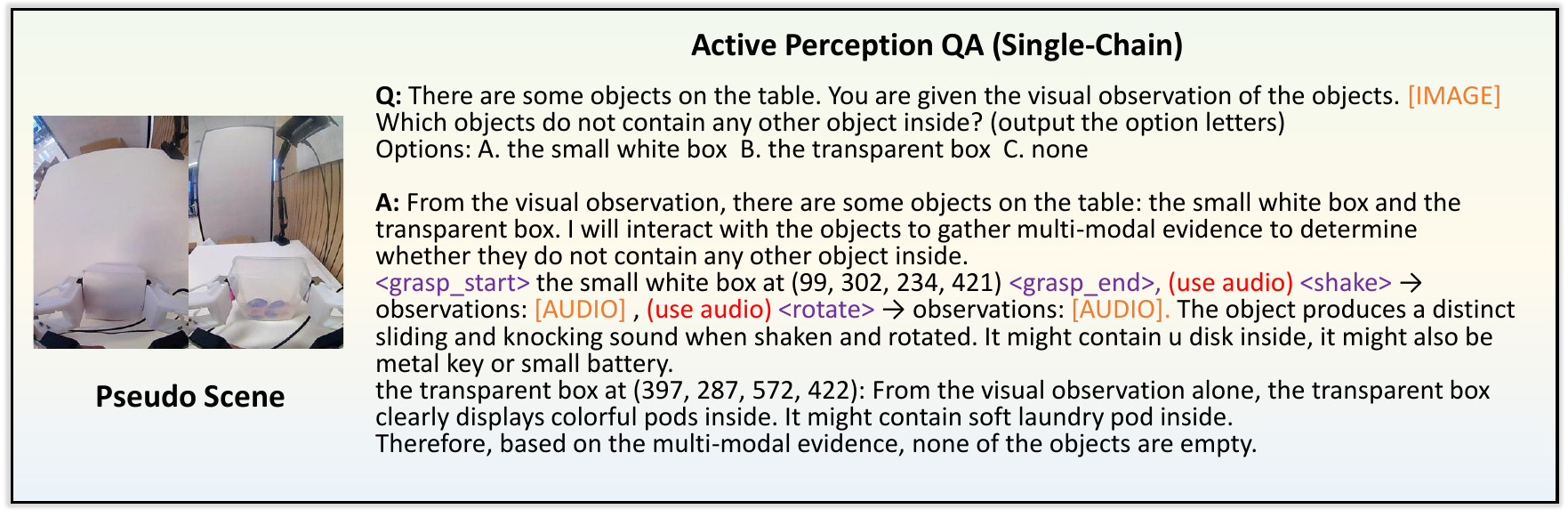}
    \caption{An example of a single-chain active perception QA sample constructed using the handheld object subset of ROMI-2K.}
    \label{fig:single qa}
\end{figure}

\begin{figure}[h]
    \centering
    \includegraphics[width=0.9\linewidth]{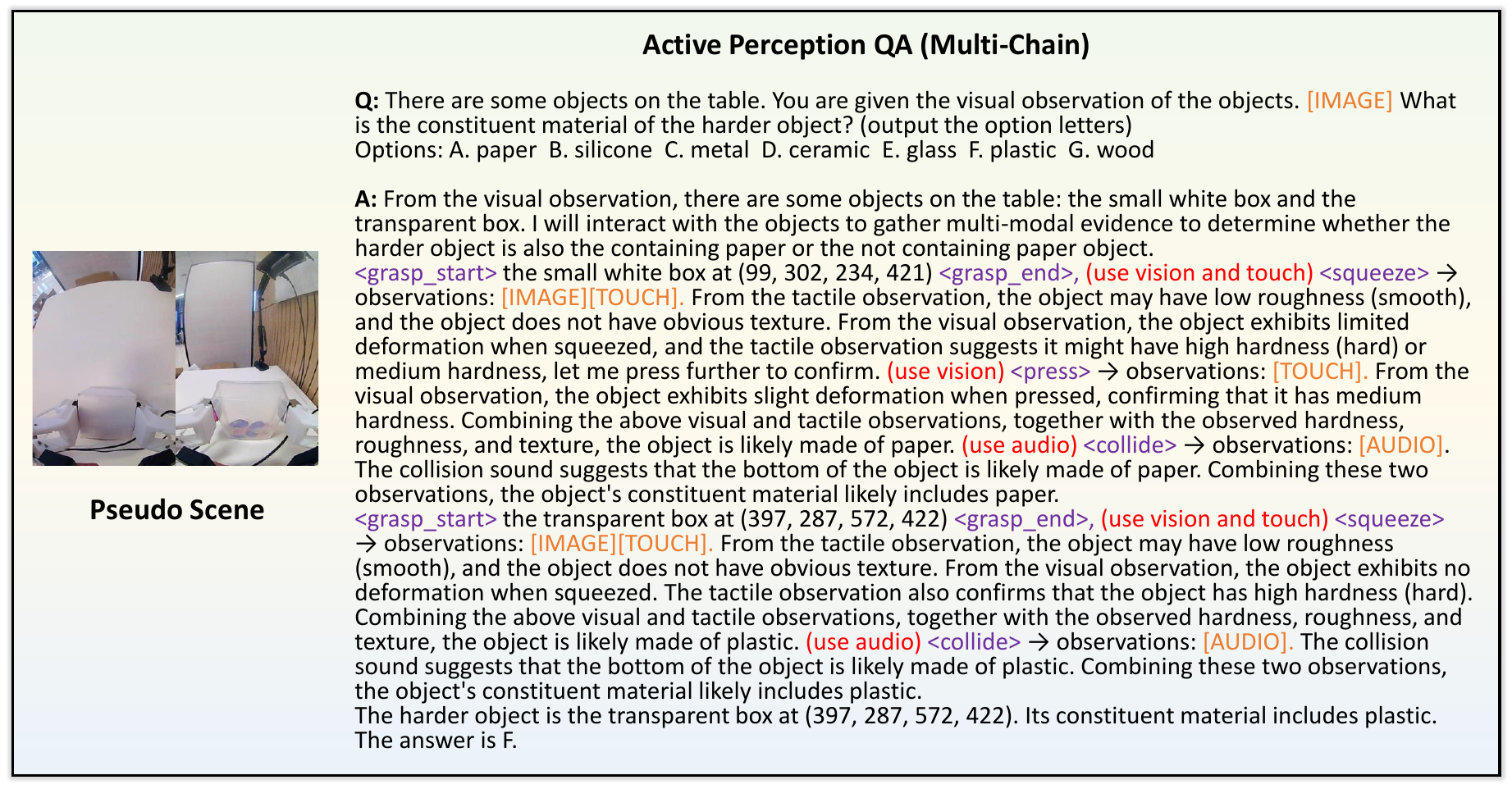}
    \caption{An example of a multi-chain active perception QA sample constructed using the handheld object subset of ROMI-2K.}
    \label{fig:multi qa}
\end{figure}

We then construct active perception QA pairs by creating pseudo-scenes from the collected handheld object data. Specifically, we crop object images from 1-3 collected handheld objects and composite them into a single scene. This process produces diverse object arrangements while preserving the visual appearance of the original objects. We construct two types of active perception tasks: single-chain and multi-chain, as shown in Fig.~\ref{fig:single qa} and~\ref{fig:multi qa}. Single-chain questions require the model to acquire evidence and determine a single target attribute, while multi-chain questions require reasoning over multiple attributes of one or more objects and may involve multiple interaction steps. The target attributes include hardness, roughness, texture, material, and inside object. For each pseudo-scene, we generate questions in multiple formats, including multiple-choice, ranking, binary judgment, and open-ended question-answering. The questions are designed such that their answers can be determined from visual observations alone or through appropriate active interactions. These pseudo-scene QA pairs provide training supervision for ROMA to identify relevant objects, select informative interactions, acquire complementary sensory evidence, and reason over the acquired observations to answer object-centric queries.

\section{Tabletop Scene Collection and Robotic System Details}
\label{app:Tabletop}

\begin{figure}[h]
    \centering
    \includegraphics[width=0.99\linewidth]{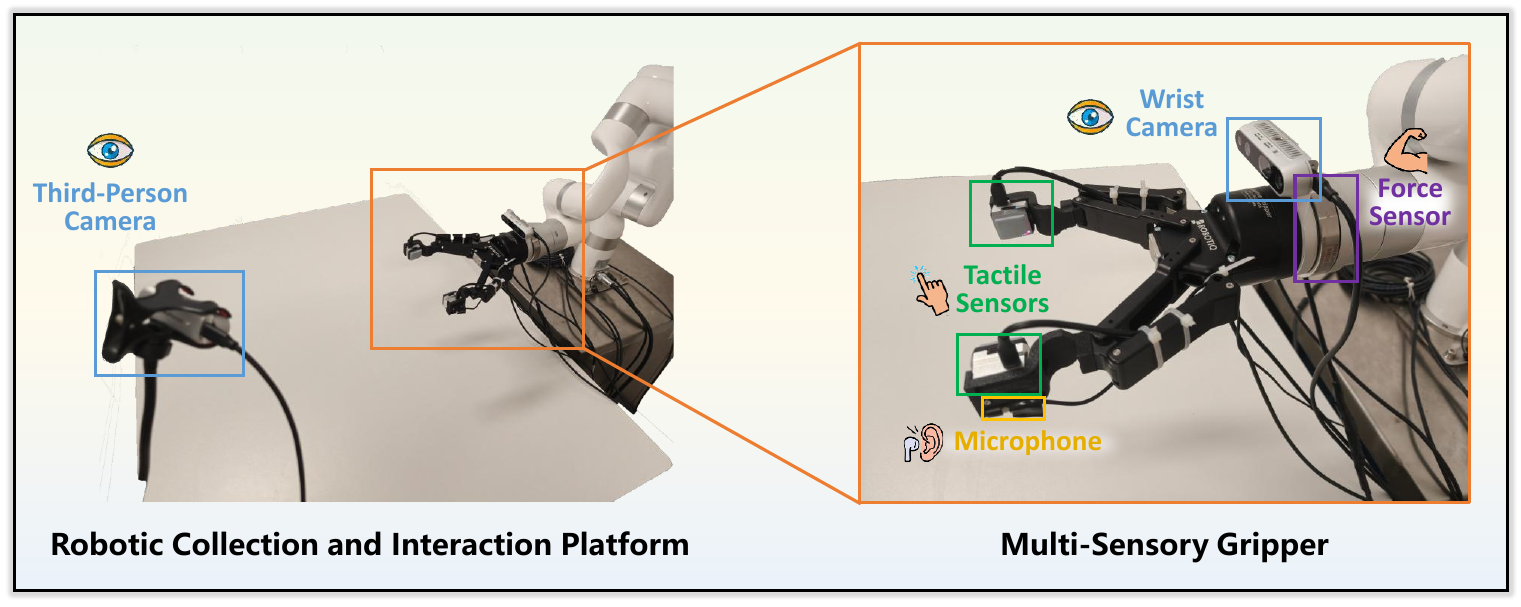}
    \caption{The illustration of the real-world robotic arm collection and interaction platform.}
    \label{fig:robotplatform}
\end{figure}
In this section, we describe the robotic system developed for ROMA. The system is designed for large-scale, multi-sensory data acquisition through physical object interaction. Given the name of an object and a corresponding visual prompt, the system autonomously detects the target from multi-view RGB-D observations, estimates its 3D geometry, plans and executes a feasible 6-DoF grasp, and performs a sequence of pre-defined physical interactions while synchronously recording visual, audio, tactile, and force feedback. Once initialized, the entire pipeline runs without manual teleoperation during inference, while human intervention and corrective adjustments are permitted during data collection to ensure safe and reliable interactions.

\subsection{Hardware}
We build the robotic data collection and interaction platform based on an xArm 6 robotic arm with a Robotiq 2F-140 gripper. We equip the gripper with two GelSight Mini tactile sensors and mount an Intel RealSense D435 on the wrist for wrist visual observations. We further deploy an Intel RealSense D435i as a fixed third-person camera, a six-axis force/torque sensor for world-frame force measurements, and a USB microphone for audio sensing. The complete hardware configuration is summarized in Tab.~\ref{tab:hardware-real}. An illustration of the real-world robotic arm collection and interaction platform is shown in Fig.~\ref{fig:robotplatform}.

\begin{table}[t]
\centering
\small
\begin{tabular}{ccc}
\hline\noalign{\smallskip}
\textbf{Hardware} & \textbf{Model} & \textbf{Role and Frequency} \\
\noalign{\smallskip}\hline\noalign{\smallskip}
Manipulator & xArm 6 & Motion Execution  \\
Gripper & Robotiq 2F-140 & Grasp + Tactile Sensor Mount \\
Wrist Camera & Intel RealSense D435 $\times$ 1 &
\makecell[c]{Wrist-view RGB Observation (10 fps) \\ Initial Scene (RGB + Depth)} \\
Third-Person Camera & Intel RealSense D435i $\times$ 1 & External RGB Observation (10 fps) \\
Force/Torque Sensor & Six-axis Force/Torque Sensor $\times$ 1 & World-Frame Force (30 Hz) \\
Tactile Sensor & GelSight Mini $\times$ 2 &Optical Tactile Collection (10 fps) \\
Microphone & USB Microphone $\times$ 1 & Audio Collection \\
\hline
\end{tabular}
\caption{Hardware configuration summary of the robotic data collection and interaction platform.}
\label{tab:hardware-real}
\end{table}

\subsection{Preparation}
Before starting each collection program, we specify the target object name and a visual prompt describing the target. The robot arm is first initialized to a predefined home pose, and the same initialization procedure is used for both data collection and inference. From the home pose, the robot captures three RGB-D observations from different viewpoints, namely the left, middle, and right views. We also record the initial force by averaging the force measurements over 2 seconds at the home pose as $F^0_{raw}$. Before the data collection of each object, we feed the middle-view RGB image together with the target object name into GroundingDINO~\citep{liu2024grounding} to obtain the target bounding box and its confidence score. To improve the reliability of the detected bounding box, we provide an interactive interface for visual inspection and correction, allowing the bounding box to be manually adjusted when necessary. During inference, the bounding box is directly obtained from the model output without an additional detection step. Next, we fuse the point clouds reconstructed from the three RGB-D observations. Each point cloud is transformed into the robot base coordinate frame using the corresponding camera extrinsics. We then apply voxel-based downsampling to reduce redundant points and remove isolated voxels that are sufficiently distant from their neighboring voxels, yielding a cleaner point cloud for subsequent grasp planning.

\subsection{Grasp Interface}
\label{app:graspmodule}

\begin{figure}[h]
    \centering
    \includegraphics[width=0.9\linewidth]{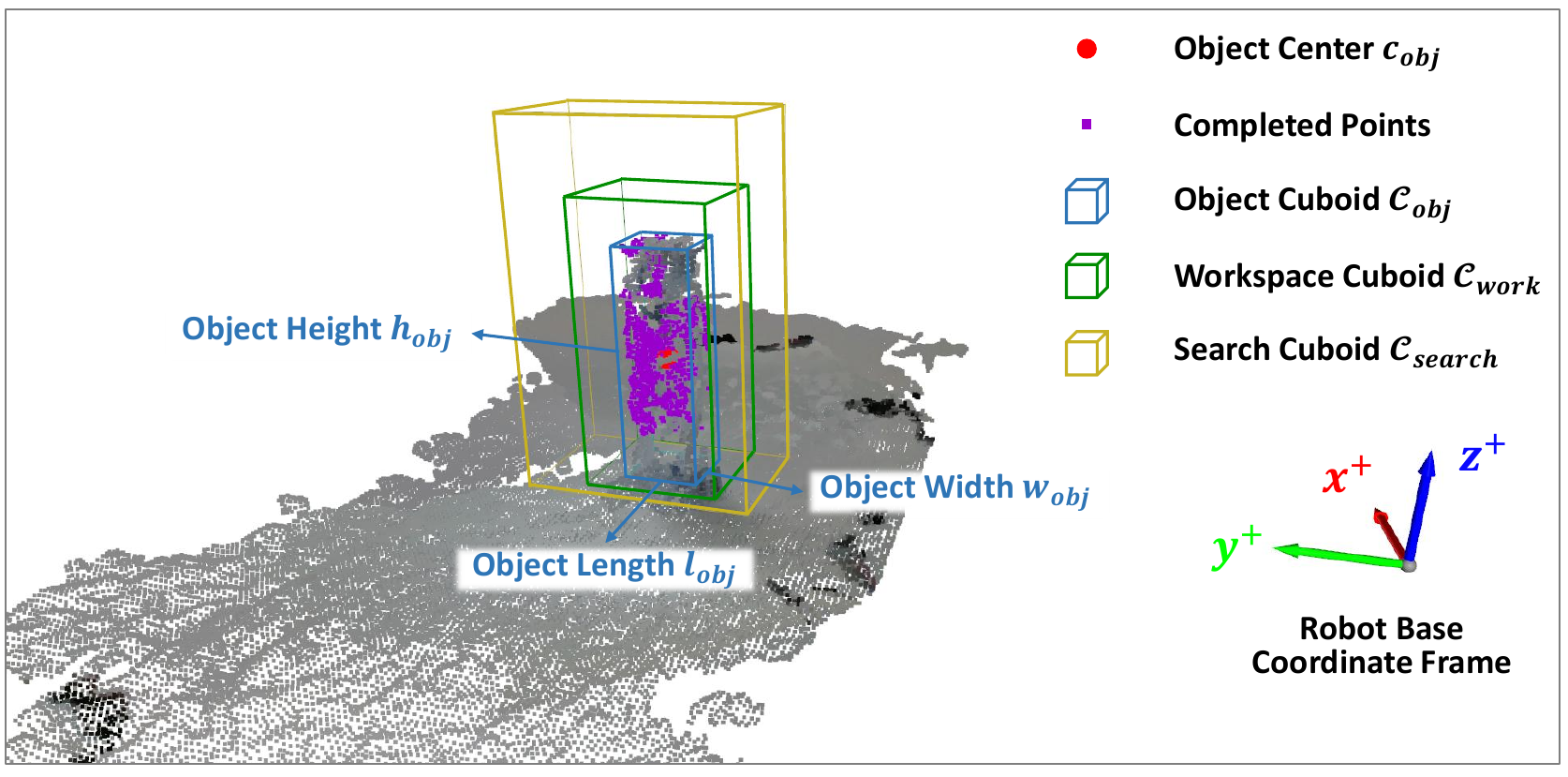}
    \caption{Visualization of the 3D point cloud constructed by the grasping module in the ROMA system, the robot base coordinate frame, and the spatial cuboids. The predicted object bounding box is first projected and uniformly expanded into a search-margin cuboid $\mathcal{C}_{\mathrm{search}}$, which defines the region for detecting and completing the object point cloud. Each object has an object center $\mathbf{c}_{\mathrm{obj}}$ and is tightly enclosed by an object cuboid $\mathcal{C}_{\mathrm{obj}}$ with an object width $w_{\mathrm{obj}}$, length $l_{\mathrm{obj}}$, and height $h_{\mathrm{obj}}$. The grasp pose for each object can only be generated within the workspace cuboid $\mathcal{C}_{\mathrm{work}}$, which is obtained by uniformly expanding the object cuboid $\mathcal{C}_{\mathrm{obj}}$ by a predefined scale factor.}
    \label{fig:cloudbase}
\end{figure}

The grasp interface converts the multi-view RGB-D observations into a feasible and robust grasp pose for subsequent object interaction. It consists of four stages: object position and size estimation, point cloud completion, grasp pose optimization, and adaptive grasping. The first two stages improve the geometric representation of the target object, while the latter two generate a reachable grasp pose and adapt the grasping force to the object.

\paragraph{Object Extent Estimation.}
We process the depth values within the bounding box using the interquartile range (IQR) method to remove depth outliers and compute the estimated object depth $d_{\mathrm{obj}}$ as the average depth of the remaining pixels. The detected bounding box is then projected into the robot base frame and proportionally enlarged to define a search-margin cuboid $\mathcal{C}_{\mathrm{search}}$. Its height is determined from $d_{\mathrm{obj}}$ and positioned above the tabletop to include the complete object point cloud while suppressing irrelevant table points.
We further remove spatial outliers from the points within $\mathcal{C}_{\mathrm{search}}$. For each coordinate axis $a\in{x,y,z}$ of the robot base frame, we estimate the corresponding object bounds by averaging the $n$ largest and $n$ smallest remaining coordinates:
\begin{equation}
a_{\max} = \frac{1}{n}\sum_{i=1}^{n}a_i^{\mathrm{top}},
\qquad
a_{\min} = \frac{1}{n}\sum_{i=1}^{n}a_i^{\mathrm{bottom}},
\end{equation}
where $a_i^{\mathrm{top}}$ and $a_i^{\mathrm{bottom}}$ denote the $n$ largest and $n$ smallest coordinates along axis $a$, respectively. Here, $x$, $y$, and $z$ denote the three coordinate axes of the robot base frame. The resulting bounds define the estimated object center
\begin{equation}
\mathbf{c}_{\mathrm{obj}}
=
\left(
\frac{x_{\max}+x_{\min}}{2},
\frac{y_{\max}+y_{\min}}{2},
\frac{z_{\max}+z_{\min}}{2}
\right)
=
(x_{\mathrm{obj}},y_{\mathrm{obj}},z_{\mathrm{obj}})
\end{equation}
and object dimensions
\begin{equation}
w_{\mathrm{obj}}=x_{\max}-x_{\min},\qquad
l_{\mathrm{obj}}=y_{\max}-y_{\min},\qquad
h_{\mathrm{obj}}=z_{\max}-z_{\min},
\end{equation}
where $w_{\mathrm{obj}}$, $l_{\mathrm{obj}}$, and $h_{\mathrm{obj}}$ denote the estimated object width, length, and height, respectively. These geometric estimates are subsequently used to refine the grasp center, gripper width, and grasp depth. Finally, we proportionally enlarge the estimated object cuboid to obtain a workspace cuboid $\mathcal{C}_{\mathrm{work}}$ for subsequent grasp pose planning.
An example is shown in Fig.~\ref{fig:cloudbase}.

\textbf{2. Point Cloud Completion.}
We employ two lightweight, symmetry-based methods to complete the object point clouds. (1) Inspired by~\citep{rumezhak2021towards}, for each cluster within the workspace, we select the top portion of the point cloud, corresponding to the highest 30\% of points along the $z$-axis by default, and use Principal Component Analysis (PCA) to estimate two principal axes. We then generate additional points by reflecting the observed points across these axes. To avoid introducing redundant or implausible points, we remove points outside the cluster's $xy$ bounding box or within 1~cm of the original point cloud based on nearest-neighbor distance. (2) For clusters with upward-facing openings, which are identified when the ratio of points within half of the estimated rim radius around the opening center falls below a predefined threshold, we fill the missing region using the following procedure. We select the top 15\% of points along the $z$-axis by default and project them onto the $xy$ plane. We then compute their convex hull to approximate the opening boundary, uniformly sample candidate points within the hull's bounding box, and retain only those lying inside the convex hull. A fixed number of valid points, 2,000 by default, are assigned the mean $z$-coordinate of the top slice and inserted into the cluster to fill the opening. Together, these two methods compensate for missing or corrupted depth measurements caused by reflective and transparent object surfaces, yielding a more complete geometric representation for the grasp planning.

\paragraph{3. Grasp Pose Optimization.}
We first generate a set of candidate grasp poses using AnyGrasp~\citep{fang2023anygrasp} from the refined point cloud.
Following the grasp representation in AnyGrasp, each candidate is parameterized by a rotation matrix $R_{\mathrm{pred}}$, a grasp center $(x_{\mathrm{center}},y_{\mathrm{center}},z_{\mathrm{center}})$, a gripper width $w_{\mathrm{pred}}$, and an approach depth $d_{\mathrm{pred}}$.
For ease of presentation, we assume that the grasp center and rotation matrix have been transformed into the robot base frame throughout the following description.
The predicted gripper tip position $\mathbf{p}_{\mathrm{tip,pred}}$ is obtained by translating the predicted grasp center by the predicted approach depth along the approach direction:
\begin{equation}
\mathbf{p}_{\mathrm{tip,pred}}
=
(x_{\mathrm{pred}},y_{\mathrm{pred}},z_{\mathrm{pred}})^{T}
=
(x_{\mathrm{center}},y_{\mathrm{center}},z_{\mathrm{center}})^{T}
+
d_{\mathrm{pred}} \cdot R_{\mathrm{pred}}[:,2],
\end{equation}
where $R_{\mathrm{pred}}[:,2]$ denotes the gripper approach direction.
In practice, the point cloud is represented in a top-down coordinate frame for grasp generation, while the resulting grasp poses are subsequently transformed into the robot base frame. 

To ensure that the generated grasp poses are compatible with the robot's workspace and feasible grasp configurations, we first filter the candidates according to their height, orientation, and workspace constraints. We consider two grasp configurations: a \emph{horizontal grasp}, where the gripper $x$-axis is within $50^\circ$ of the base $z$-axis and the gripper $z$-axis is within $50^\circ$ of the base $y$-axis, and a \emph{vertical grasp}, where the gripper $z$-axis is within $50^\circ$ of the base $z$-axis and the gripper $x$-axis is within $50^\circ$ of the base $y$-axis. Candidates with $z<120$~mm are retained for horizontal grasps, while those with $z>55$~mm are retained for vertical grasps. We further constrain the candidates to the valid workspace, with the grasp height and lateral $y$-coordinate restricted.
For the remaining candidates, we refine the grasp geometry using the estimated object geometry. The gripper width $w_{\mathrm{pred}}$ predicted by AnyGrasp is then adjusted according to the estimated object width $w_{\mathrm{obj}}$ as
\begin{equation}
w_\mathrm{final} =
0.8 \cdot k_w \cdot  w_{\mathrm{pred}}
+
(1-k_w) \cdot  w_{\mathrm{obj}},
\end{equation}
where $w_\mathrm{final}$ denotes the adjusted gripper opening width and $k_w$ controls the relative contribution of the predicted and point-cloud-based widths. The factor $0.8$ provides a conservative reduction of the predicted opening width. This adjustment makes the gripper opening better match the estimated object size, reducing excessive opening and improving grasp stability, particularly for relatively soft objects. 

We also refine the $x$ and $y$ coordinates of the gripper tip position point $\mathbf{p}_{\mathrm{tip,pred}}$ predicted by AnyGrasp using the estimated object center $\mathbf{c}_{\mathrm{obj}}$. Specifically, we obtain the $x$ and $y$ coordinates of the final grasp center $(x_{\mathrm{final}},y_{\mathrm{final}})$ as:
\begin{equation}
x_{\mathrm{final}} =
k_c \cdot  x_{\mathrm{pred}}
+
(1-k_c) \cdot x_{\mathrm{obj}},
\qquad
y_{\mathrm{final}} =
k_c  \cdot y_{\mathrm{pred}}
+
(1-k_c) \cdot y_{\mathrm{obj}},
\end{equation}
where $(x_{\mathrm{obj}},y_{\mathrm{obj}})$ denotes the corresponding coordinates of $\mathbf{c}_{\mathrm{obj}}$, and $k_c$ controls the relative contribution of the predicted grasp center and the estimated object center. Only the $x$ and $y$ coordinates are refined, while the $z$ coordinate remains unchanged. This adjustment moves the grasp center closer to the object center, reducing the risk of the grasp pose being biased toward one side of the object and causing the object to be pushed away by unilateral contact during gripper closure.

We further refine the grasp depth $d_{\mathrm{pred}}$ for \textbf{horizontal grasps} using the estimated object geometry. Specifically, we combine the predicted grasp depth $d_{\mathrm{pred}}$ with an object-dependent reference depth $l_{\mathrm{obj}}$:
\begin{equation}
\alpha_\mathrm{horizontal} =
\frac{k_d \cdot  d_{\mathrm{pred}}}
{d_{\mathrm{pred}}+0.7 \cdot l_{\mathrm{obj}}},
\end{equation}
\begin{equation}
d_\mathrm{horizontal} =
\alpha_\mathrm{horizontal} \cdot  d_{\mathrm{pred}}
+
(1-\alpha_\mathrm{horizontal}) \cdot (0.7 \cdot l_{\mathrm{obj}}),
\end{equation}
where $l_{\mathrm{obj}}$ denotes the estimated object length, $d_{\mathrm{pred}}$ denotes the predicted grasp depth, $d_\mathrm{horizontal}$ denotes the refined grasp depth, and $k_d$ controls the contribution of the predicted depth. The term $0.7 \cdot l_{\mathrm{obj}}$ provides a geometry-based reference depth, allowing the grasp depth to adapt to the object's size and placing the grasp closer to the object center for longer objects.
For \textbf{vertical grasps}, we similarly refine the grasp depth based on the object height:
\begin{equation}
\alpha_\mathrm{vertical} =
\frac{k_d \cdot  d_{\mathrm{pred}}}
{d_{\mathrm{pred}}+(0.7 \cdot h_{\mathrm{obj}})},
\end{equation}
\begin{equation}
d_\mathrm{vertical} =
\alpha_\mathrm{vertical}  \cdot d_{\mathrm{pred}}
+
(1-\alpha_\mathrm{vertical}) \cdot (0.7 \cdot h_{\mathrm{obj}}),
\end{equation}
where $d_\mathrm{vertical}$ denotes the adjusted grasp depth for vertical grasps and $k_d$ is the same parameter that controls the contribution of the predicted grasp depth.
Since we use the Robotiq 2F-140 gripper, which employs a four-bar linkage mechanism rather than the parallel-jaw gripper used in AnyGrasp, the resulting grasp depth is further compensated for the circular motion of the gripper fingers. We assume the fingertips follow a circular trajectory with radius $r_{\mathrm{gripper}}=100$~mm (for Robotiq 2F-140), and the compensation distance is
\begin{equation}
\Delta_{\mathrm{gripper}} =
r_{\mathrm{gripper}}-\sqrt{r_{\mathrm{gripper}}^2-\left(\frac{w_\mathrm{final}}{2}\right)^2},
\end{equation}
where $\Delta_{\mathrm{gripper}}$ denotes the additional depth required to account for the finger trajectory and $r_{\mathrm{gripper}}$ is the effective radius of the gripper-finger motion. The grasp depth offset is then
\begin{equation}
d_{\mathrm{offset}} = d_{h/v}+\Delta_{\mathrm{gripper}},
\end{equation}
where $d_{h/v} \in \{d_\mathrm{horizontal},d_\mathrm{vertical}\}$ denotes $d_\mathrm{horizontal}$ or $d_\mathrm{vertical}$ according to the selected grasp configuration. Finally, we clip the grasp depth offset $d_{\mathrm{offset}}$ and the $z$ coordinate of the grasp center $z_{\mathrm{pred}}$ to improving grasp stability.

We further refine the grasp orientation $R_{\mathrm{pred}}$ to adapt the AnyGrasp pose to the kinematic constraints of our robot. For horizontal grasps, we first rotate the gripper around its local $z$-axis to make the gripper $y$-axis approximately horizontal while preserving the approach direction. If the gripper $z$-axis points upward, we further rotate the gripper around its local $y$-axis to bring the $z$-axis closer to the base $xy$ plane, reducing the risk of the robot elbow contacting the tabletop. For vertical grasps, we first rotate the gripper around its local $x$-axis to make the gripper $y$-axis approximately horizontal. If the gripper $z$-axis points backward, we further rotate it around its local $y$-axis to bring the $z$-axis closer to the base $xz$ plane, avoiding excessive forward extension of the robot arm. The resulting rotation matrix is denoted as $R_{\mathrm{final}}$ and is used together with the refined grasp center and approach depth to construct the final grasp pose.

\begin{figure}[t]
    \centering
    \includegraphics[width=\linewidth]{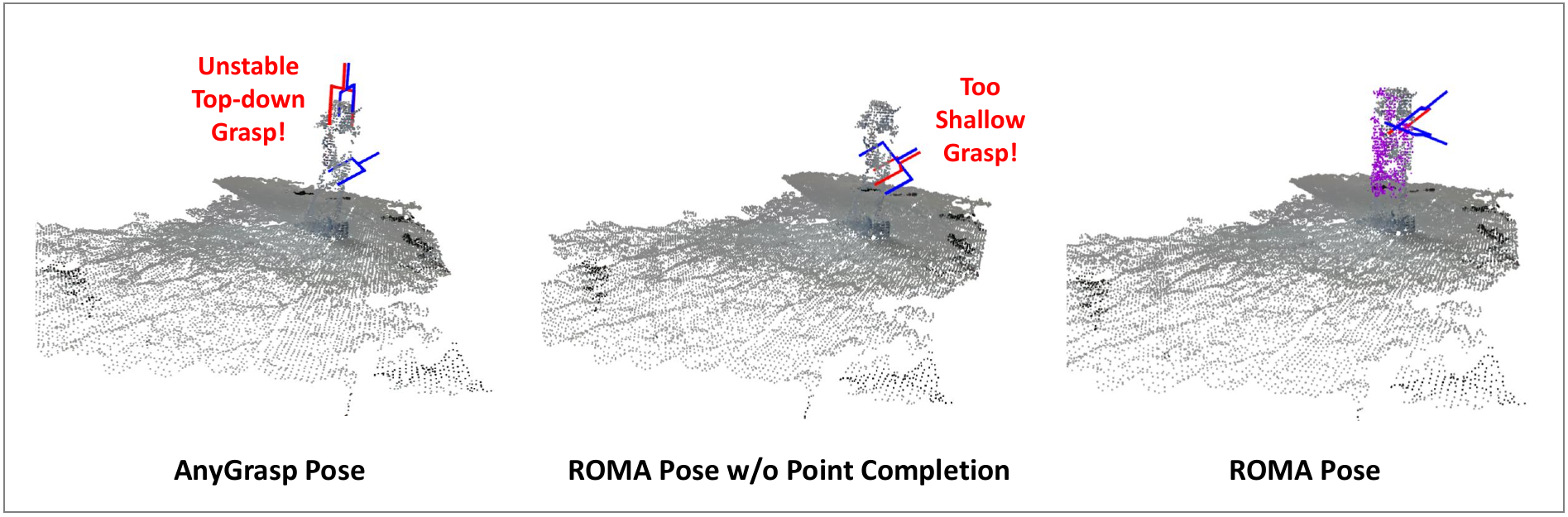}
    \caption{Visualization of the grasp poses predicted by AnyGrasp, ROMA without point cloud completion, and ROMA with point cloud completion. For each grasping module, we visualize the top-3 predicted grasp poses and highlight the top-1 grasp pose to be executed in red. The original AnyGrasp predictions often fail to provide sufficiently stable grasps for the robot to perform a sequence of interactions, particularly more vigorous actions such as shaking and rotating. Without point cloud completion, the predicted grasp poses are affected by gaps in the observed point cloud and often result in overly shallow grasps, causing objects to slip or fall. With point cloud completion, the full ROMA grasping module produces more stable and reliable grasp poses.}
    \label{fig:romagrasp}
\end{figure}

After refining the gripper width, grasp center, approach depth and grasp orientation, we construct the final grasp pose. The refined reference position is
$
(x_{\mathrm{final}},
y_{\mathrm{final}},
z_{\mathrm{pred}})^T$
where only the $x$ and $y$ coordinates are refined, while the $z$ coordinate remains unchanged. The final gripper tip position $\mathbf{p}_{\mathrm{tip,final}}$ is then computed as
\begin{equation}
\mathbf{p}_{\mathrm{tip,final}}
=
(x_{\mathrm{final}},
y_{\mathrm{final}},
z_{\mathrm{pred}})^T
+
d_{\mathrm{offset}} \cdot R_{\mathrm{final}}[:,2],
\end{equation}
where $R_{\mathrm{final}}$ denotes the refined gripper orientation and $d_{\mathrm{offset}}$ is the compensated approach depth. The resulting grasp is therefore executed with the refined width $w_{\mathrm{final}}$, orientation $R_{\mathrm{final}}$, grasp center $
(x_{\mathrm{final}},
y_{\mathrm{final}},
z_{\mathrm{pred}})^T$, and tip position $\mathbf{p}_{\mathrm{tip,final}}$. We rank the valid candidates by their AnyGrasp confidence and select the highest-confidence candidate for execution. A comparison between the grasp poses predicted by ROMA and the corresponding grasp poses predicted by the original AnyGrasp is shown in Fig.~\ref{fig:romagrasp}.

\textbf{4. Grasping.} After moving to the refined grasp pose, we use a two-stage grasping strategy to adapt the gripping force to the target object. First, the gripper closes with a low contact force of 1~N and stops upon contacting the object. We then measure the current gripper width $w_{\mathrm{current}}$ and command the gripper to close by an additional approximately 25~mm. At this position, we measure the current gripping force $F_{\mathrm{max}}^{g}$, which reflects the resistance encountered during closure. We use this measurement to determine the final gripping force as
\begin{equation}
F^g_{\mathrm{final}}
=
k_g \cdot F_{\mathrm{max}}^{g},
\end{equation}
where $k_g$ is a scaling factor (default $k_g=2.0$). The gripper then applies $F^g_{\mathrm{final}}$ to complete the grasp. This two-stage strategy allows the gripping force to adapt to the object's size and resistance without requiring object-specific force tuning.

\subsection{Interaction and Feedback}
\label{sec:interaction_feedback}

With reliable grasp poses established, ROMA further collects multi-sensory feedback during the same six pre-defined interactions in Sec.~\ref{sec:dataset}. After grasping, the robot arm returns to a predefined home pose, where the tool $z$-axis is leveled to provide a stable configuration for subsequent interactions. During each interaction, we synchronously record data from the left and right GelSight sensors, wrist and third-person RealSense cameras, USB microphone, and six-axis force/torque sensor, with timestamps for temporal alignment across modalities. The six atomic interactions are defined as follows:
\begin{itemize}
\item \textbf{\textit{Lift}:} The grasped object is first held at the home pose for 4 seconds to reduce transient effects. The arm then raises the object by 50~mm and holds it at a fixed height. We use the average force over the entire lift trajectory as gravity-related evidence, providing a robust cue for comparing the relative weight of objects.

\item \textbf{\textit{Shake}:} The robotic arm drives the grasped object through seven rapid reciprocating cycles along the $x$-axis, generating dynamic multi-sensory responses that can reveal information about the object's internal contents.

\item \textbf{\textit{Rotate}:} The object is rotated $180^\circ$ around its grasp axis, exposing different surface regions and providing complementary visual observations. The resulting motion can also produce acoustic cues about internal contents.

\item \textbf{\textit{Squeeze}:} The gripper attempts to reduce its grasping width by 8~mm, applying additional contact force to the object surface to capture multi-sensory responses under increased normal load.

\item \textbf{\textit{Collide}:} The arm rapidly taps the grasped object against the table four times in succession, producing impact-related multi-sensory responses that provide cues about the object's physical properties.

\item \textbf{\textit{Press}:} The arm moves the grasped object downward by 15~mm against the table at a controlled speed. The resulting multi-sensory response provides information about the object's compliance.
\end{itemize}

For each interaction, the recorded signals are processed into modality-specific feedback that can be provided to ROMA. Specifically, we encode one wrist-camera image, one cropped audio segment, and two optical tactile images from each interaction as optional multi-sensory inputs, with one tactile image captured at the beginning of the interaction and the other captured during the interaction. The force measurements require additional processing because the six-axis force/torque sensor is mounted at the wrist, and its raw measurements are expressed in the sensor coordinate frame, which changes with the gripper pose. Directly using these measurements would therefore make the same physical force appear differently across interactions. Therefore, we first transform the raw force data into the robot base frame and compensate for the force exerted by the empty gripper:
\begin{equation}
 F_{trans} = R_{\mathrm{B}\leftarrow \mathrm{G}}R_{\mathrm{G}\leftarrow \mathrm{S}}(F_{raw} - F^0_{raw}),
\end{equation}
where $F_{\mathrm{raw}}$ denotes the raw force data, $F_{\mathrm{raw}}^{0}$ denotes the force measured with the empty gripper, and $R_{\mathrm{G}\leftarrow\mathrm{S}}$ and $R_{\mathrm{B}\leftarrow\mathrm{G}}$ denote the rotation matrices from the sensor frame to the gripper frame and from the gripper frame to the robot base frame, respectively. The resulting $F_{\mathrm{trans}}$ represents the force exerted by the grasped object in the base coordinate frame, providing a consistent representation across different gripper poses. 
We then average the force along the gravity direction during each interaction and provide this value to the model as textual evidence. Since force measurements still contain some noise and error, we focus on the gravity-direction component, which provides a more reliable cue for weight-related reasoning, and leave the other force components for future exploration.
We also record images from a third-person camera, but do not use them during model training or inference, as we aim to integrate all sensors used by the system onto the robotic arm to improve overall system integration.

\subsection{Annotation}
\begin{figure}[h]
    \centering
    \includegraphics[width=0.9\linewidth]{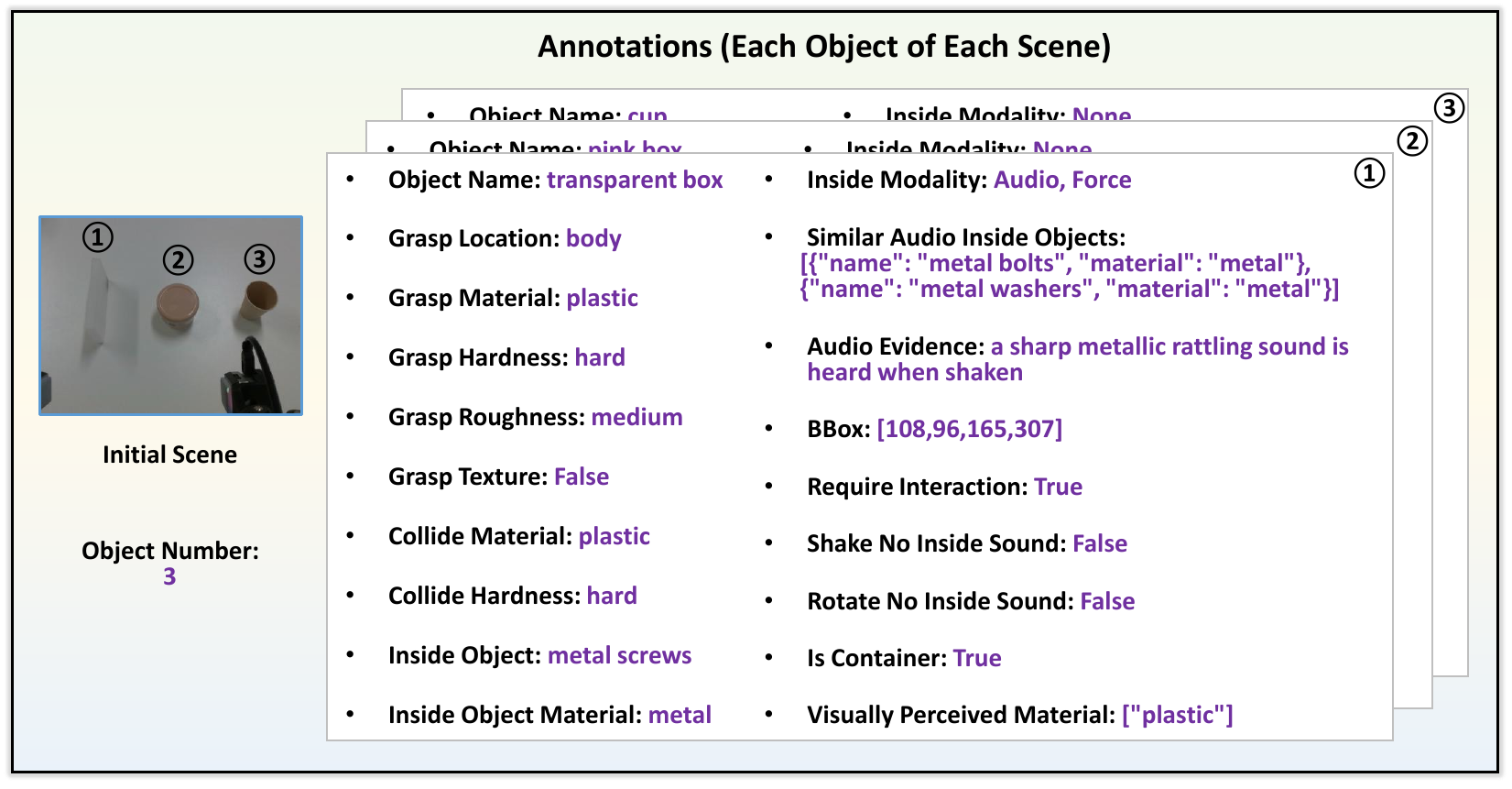}
    \caption{An example of a multi-sensory data sample and its annotations from the tabletop scene subset of ROMI-2K.}
    \label{fig:robot anno}
\end{figure}
Similar to the handheld object subset, we annotate each collected object with its name and the same physical attributes mentioned in Appendix~\ref{app:hand anno}. For each tabletop scene, every object instance is independently annotated with its object identity and corresponding physical attributes. 
In addition, since the tabletop platform is equipped with a force sensor, we can use force feedback as an additional modality for inferring whether a container contains an object. Specifically, for objects with potentially heavy contents, we compare the measured force with that of the corresponding empty container. If the force reading is noticeably higher than that of the empty container, we mark force as an available modality for determining the presence of an internal object.
We first use Gemini 3.5 Flash to generate annotations for each object in the scene, and then manually inspect and correct the annotations on an object-by-object basis to ensure their accuracy. These object-level annotations are associated with the individual objects in each scene and serve as ground truth for constructing scene-level active perception QA pairs and evaluating the model's active perception ability. An example is shown in Fig.~\ref{fig:robot anno}.

\subsection{QA Construction}
\begin{figure}[h]
    \centering
    \includegraphics[width=0.9\linewidth]{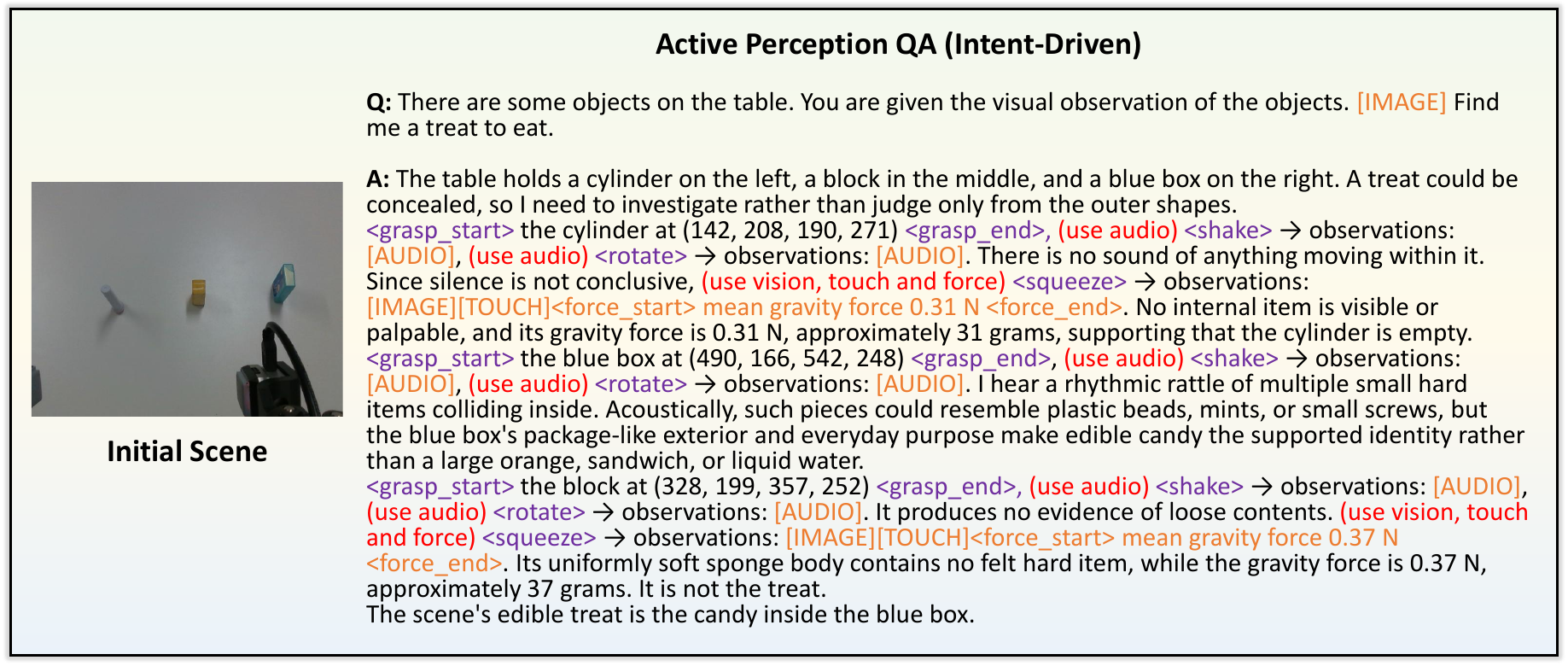}
    \caption{An example of an intent-driven active perception QA sample constructed using the tabletop scene subset of ROMI-2K.}
    \label{fig:intentqa}
\end{figure}
Similar to Appendix~\ref{app:hand qa}, we construct scene-level active perception QA pairs based on the object-level annotations of each tabletop scene. Specifically, we use the annotated object identities and physical attributes to generate questions that require the model to identify relevant objects, acquire the necessary sensory evidence through physical interactions, and reason over the resulting observations. We construct QA pairs in four formats, including multiple-choice, fill-in-the-blank, true-or-false, and open-ended question-answering. The questions cover different object-centric attributes and are designed to require either direct visual reasoning or active acquisition of complementary sensory information. 
Following the design of the handheld subset, we include both single-chain and multi-chain active perception tasks, where single-chain tasks require the model to acquire evidence for determining a single target attribute, while multi-chain tasks require reasoning over multiple attributes of one or more objects and may involve multiple interaction steps. In addition, we introduce an intent-driven task that requires the model to infer the user's underlying information need and actively determine which objects to interact with, which interactions to perform, and which sensory modalities to acquire. For the intent-driven task, we provide GPT-5.4 with the complete object-level annotations of a scene, together with descriptions of the available actions and sensory modalities. GPT-5.4 then generates QA pairs by specifying an implicit information-seeking intent that can be resolved through appropriate active perception. This construction allows the resulting questions to go beyond explicitly stated attribute queries and requires the model to infer the relevant object, interaction, and modality from the user's intent. An example of the intent-driven task is shown in Fig.~\ref{fig:intentqa}.
For all tasks involving hardness, roughness, and weight, we avoid ambiguous questions that require judging whether an individual object is simply hard, rough, or heavy. Instead, we formulate comparative questions across objects, such as identifying the hardest, roughest, or heaviest object. We further ensure a sufficiently clear difference between the correct answers and the distractors, thereby improving the quality and consistency of the resulting QA tasks.
For training, we use QA pairs constructed from the training subset of tabletop scenes to provide supervision for scene-level active perception, including object selection, interaction selection, and multi-sensory reasoning. For evaluation, we construct a separate set of QA pairs from a held-out subset of tabletop objects. None of the objects in this evaluation subset overlap with the objects used in either the tabletop training data or the handheld object data. This object-disjoint split enables us to evaluate the model's ability to generalize active perception to previously unseen objects in real-world tabletop scenes.

\clearpage
\section{Dataset Statistics}
\label{app:dataset}
\begin{figure}[h]
    \centering
    \includegraphics[width=0.99\linewidth]{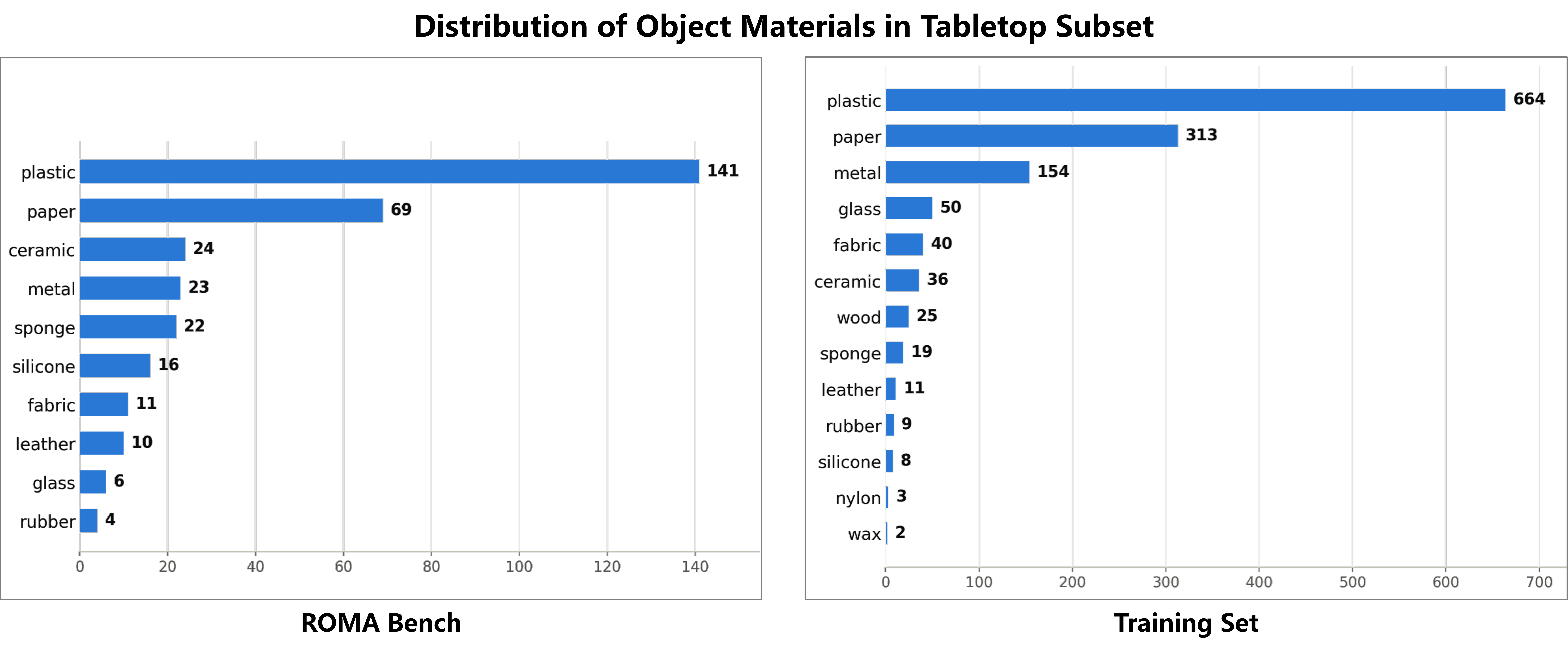}
    \caption{Statistics of object materials in the tabletop scene subset of ROMI-2K.}
    \label{fig:table-material}
\end{figure}

\begin{figure}[h]
    \centering
    \includegraphics[width=0.9\linewidth]{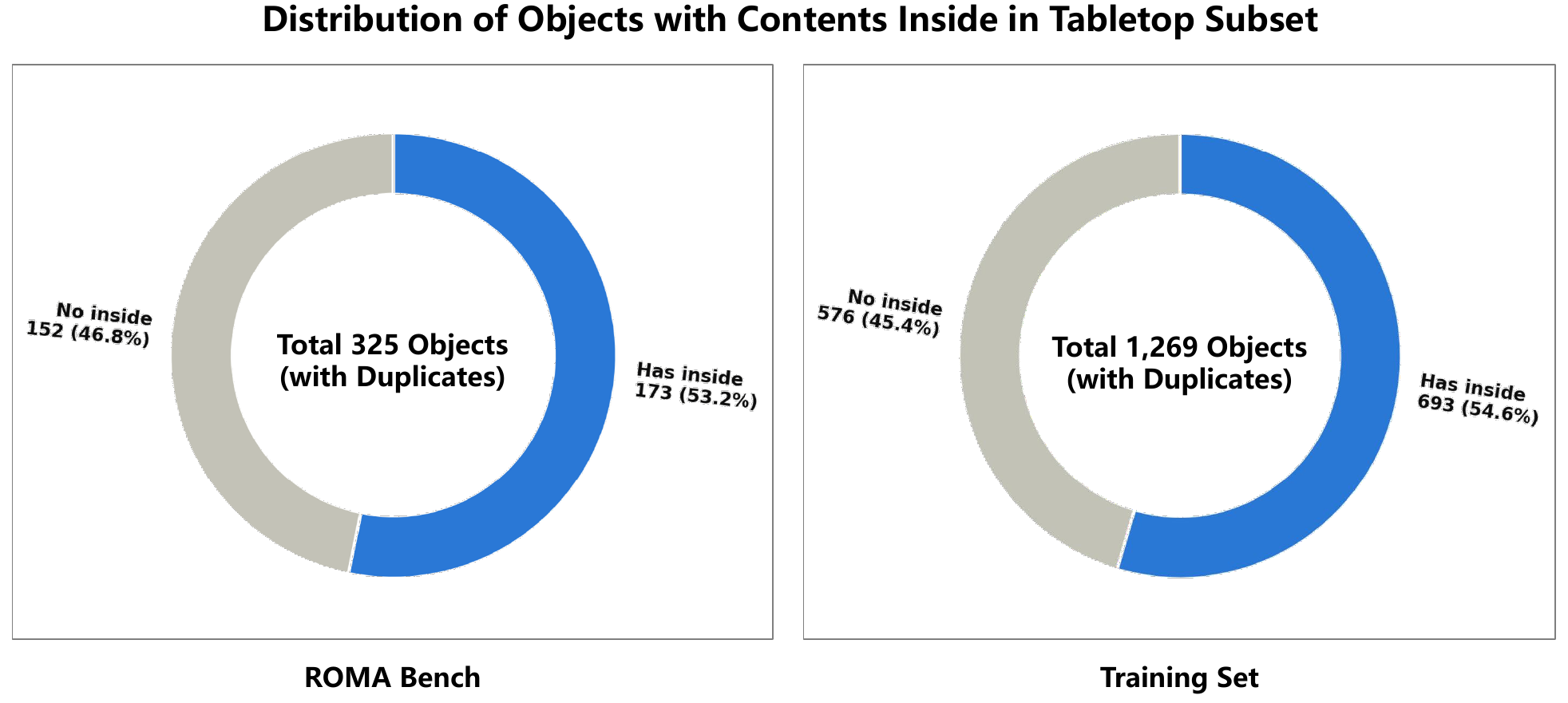}
    \caption{Statistics of objects with inside contents in the tabletop scene subset of ROMI-2K.}
    \label{fig:table-inside}
\end{figure}

\begin{figure}[h]
    \centering
    \includegraphics[width=0.9\linewidth]{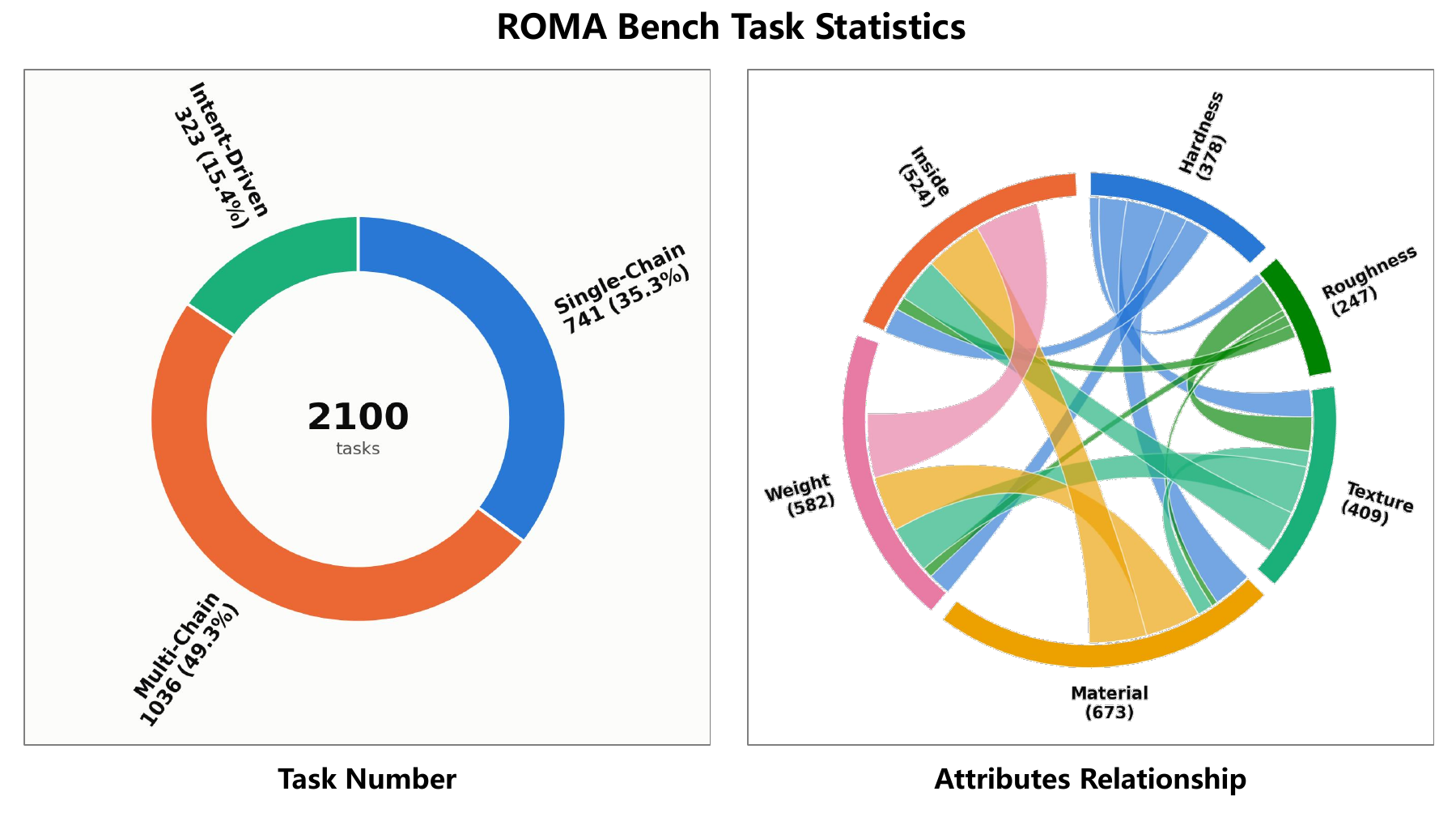}
    \caption{Number of tasks by type and the relationship between target attributes in Single-Chain and Multi-Chain QA pairs in ROMA Bench.}
    \label{fig:bench-xiantu}
\end{figure}

\begin{figure}[h!]
    \centering
    \includegraphics[width=0.9\linewidth]{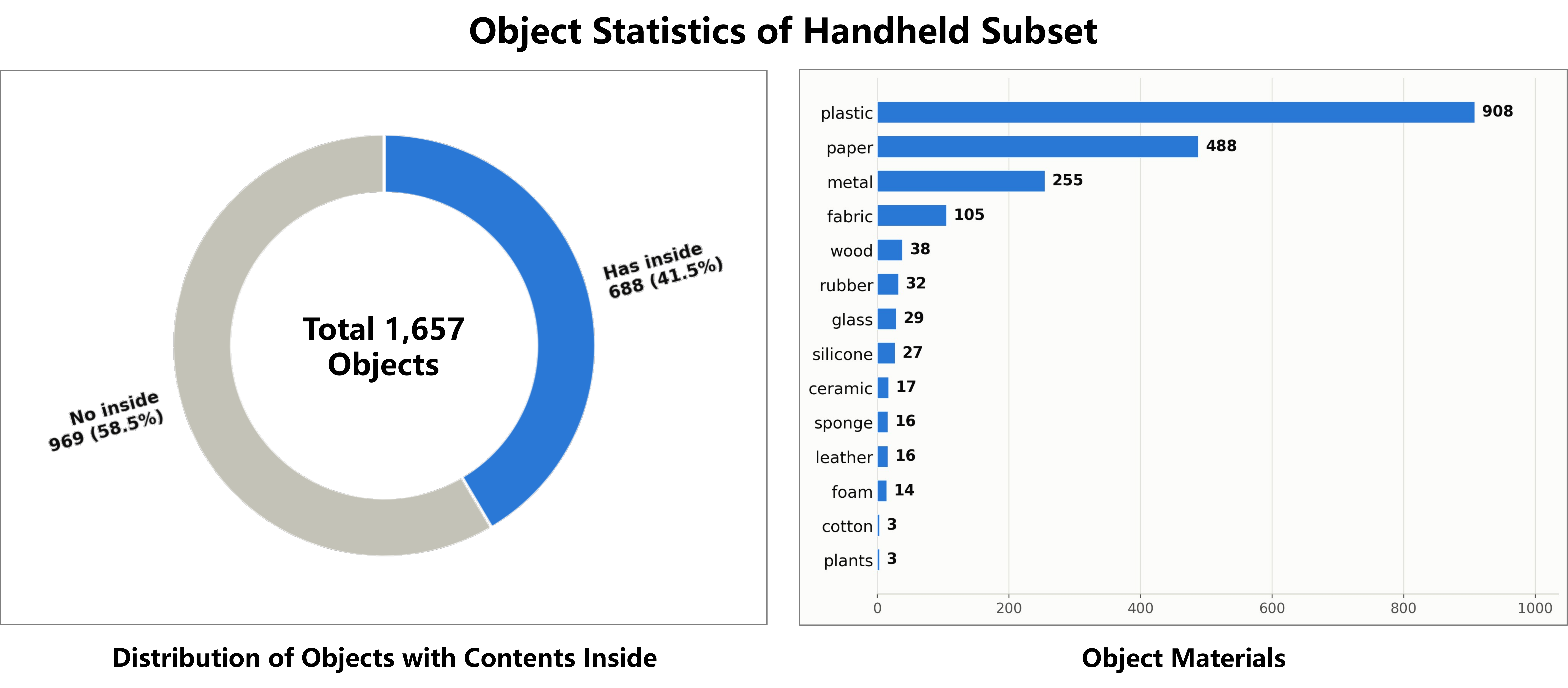}
    \caption{Statistics of the handheld object subset in ROMI-2K. Left: distribution of objects with and without contents. Right: distribution of the top 14 material categories.}
    \label{fig:hand-stat}
\end{figure}

In this section, we first provide a detailed statistical analysis of the tabletop scene subset (including both the training set and ROMA Bench) of ROMI-2K. We provide statistics on the object materials, the presence of contents within objects, and the task statistics of ROMA Bench for the training set and ROMA Bench of the tabletop scene subset, as shown in Fig.~\ref{fig:table-material}, \ref{fig:table-inside}, and \ref{fig:bench-xiantu}, respectively.
As shown in these figures, the 1,269 object instances in the training set and the 325 object instances in ROMA Bench exhibit diverse material compositions, covering 13 different material categories, with plastic, metal, and paper being the most prevalent. The presence of contents is also relatively balanced, with objects containing contents and empty objects each accounting for approximately half of the objects. Such diversity in object materials and contents provides broad combinations of physical properties and interaction outcomes, enabling ROMA to learn more varied cross-modal sensory cues and improving the coverage of active perception scenarios.
In addition, the Multi-Chain tasks in ROMA Bench exhibit rich and relatively balanced relationships among different target attributes, requiring the model to acquire and integrate complementary sensory evidence across multiple attributes. This diverse attribute composition provides a more comprehensive evaluation of the model's ability to perform multi-step active perception and reason over multiple sensory observations. It is worth noting that the same object (or object-content combination) may appear in multiple scenes. We do not deduplicate or distinguish such repeated instances here.

We further analyze the statistics of the handheld object subset, as shown in Fig.~\ref{fig:hand-stat}. The subset contains 1,657 object-content combinations and covers 20 different material categories. As shown in the figure, the material distribution is diverse, with the top 14 categories accounting for the majority of the collected objects. In terms of object contents, 41.5\% of the object combinations contain contents, while the remaining objects are empty. The diverse material and content configurations provide broad coverage of object properties and interaction outcomes, complementing the tabletop subset with additional object-level interaction diversity.

\clearpage
\section{Implementation Details}
\label{app:Implementation}
\begin{table}[h]
\centering
\small
\tabcolsep=0.7cm
\begin{tabular}{ll}
\hline\noalign{\smallskip}
\textbf{Hyperparameter} & \textbf{Setting} \\
\noalign{\smallskip}\hline\noalign{\smallskip}
\multicolumn{2}{l}{\textbf{Stage 1: Audio Alignment}} \\
Trainable Components & Audio Branch \\
Frozen Components & LLM + Vision Branch + Tactile Branch \\
Precision & BF16 \\
Effective Batch Size & 96 \\
Learning Rate & $1\times10^{-4}$ \\
Weight Decay & 0 \\
Warmup Ratio & 0.03 \\
LR Scheduler & Cosine \\
Epochs & 1 \\
Max Sequence Length & 2,048 \\
Gradient Clipping & 4.0 \\
\noalign{\smallskip}\hline\noalign{\smallskip}
\multicolumn{2}{l}{\textbf{Stage 1: Tactile Alignment}} \\
Trainable Components & Tactile Adapter \\
Frozen Components & LLM + Vision Branch + Audio Branch + Tactile Encoder \\
Precision & BF16 \\
Effective Batch Size & 48 \\
Learning Rate & $1\times10^{-4}$ \\
Weight Decay & 0 \\
Warmup Ratio & 0.03 \\
LR Scheduler & Cosine \\
Epochs & 1 \\
Max Sequence Length & 2,048 \\
Gradient Clipping & 4.0 \\
\noalign{\smallskip}\hline\noalign{\smallskip}
\multicolumn{2}{l}{\textbf{Stage 2: Multi-Sensory SFT}} \\
Trainable Components & LLM \\
Frozen Components & Audio Branch + Tactile Branch + Vision Branch \\
Precision & BF16 \\
Effective Batch Size & 24 \\
Learning Rate & $1\times10^{-5}$ \\
Weight Decay & 0 \\
Warmup Ratio & 0.03 \\
LR Scheduler & Cosine \\
Epochs & 1 \\
Max Sequence Length & 8,192 \\
Dynamic Scene Sampling Factor $\alpha$ & 2.5 \\
\noalign{\smallskip}\hline
\end{tabular}
\caption{Training hyperparameters for the two-stage alignment and supervised fine-tuning of ROMA-7B. The ``Branch'' refers to the encoder and adapter.}
\label{tab:training-hyperparams}
\end{table}

\begin{figure}[ht]
    \centering
    \includegraphics[width=0.9\linewidth]{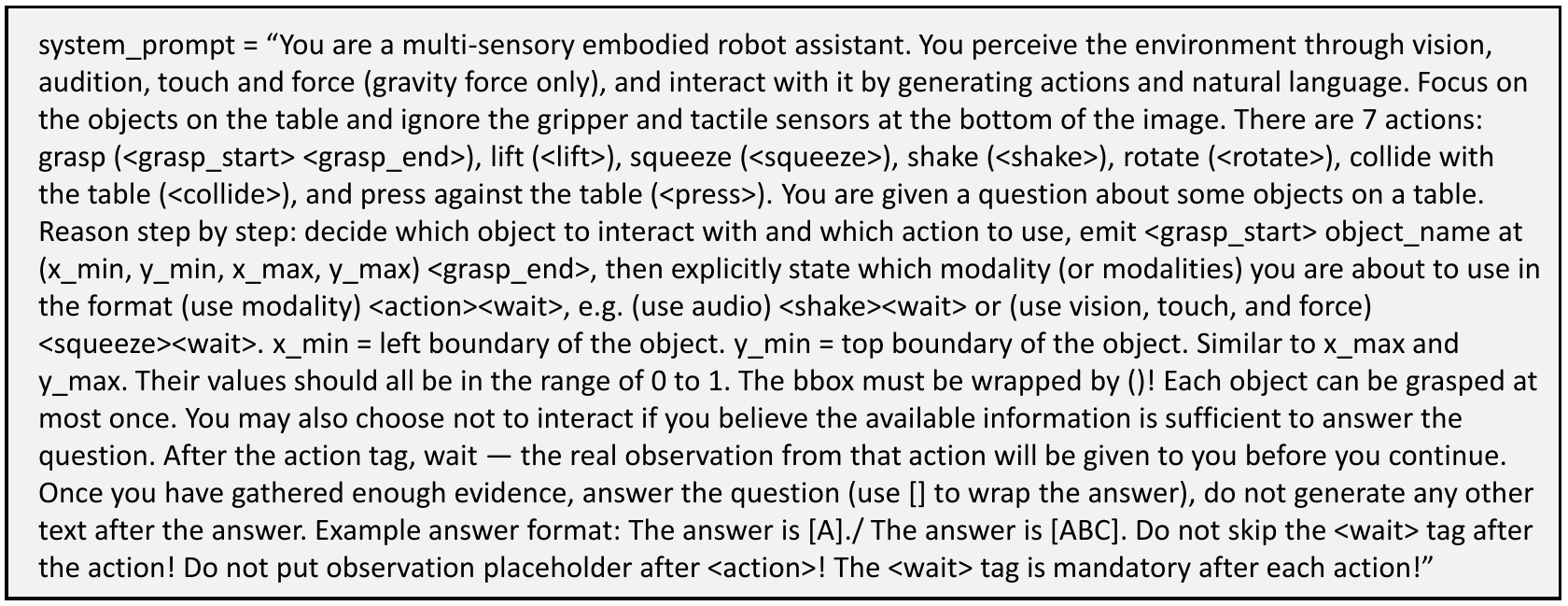}
    \caption{The system prompt for Gemini 3.5 Flash in ROMA Bench active perception evaluation.}
    \label{fig:gemini-sys}
\end{figure}

\begin{figure}[ht]
    \centering
    \includegraphics[width=0.9\linewidth]{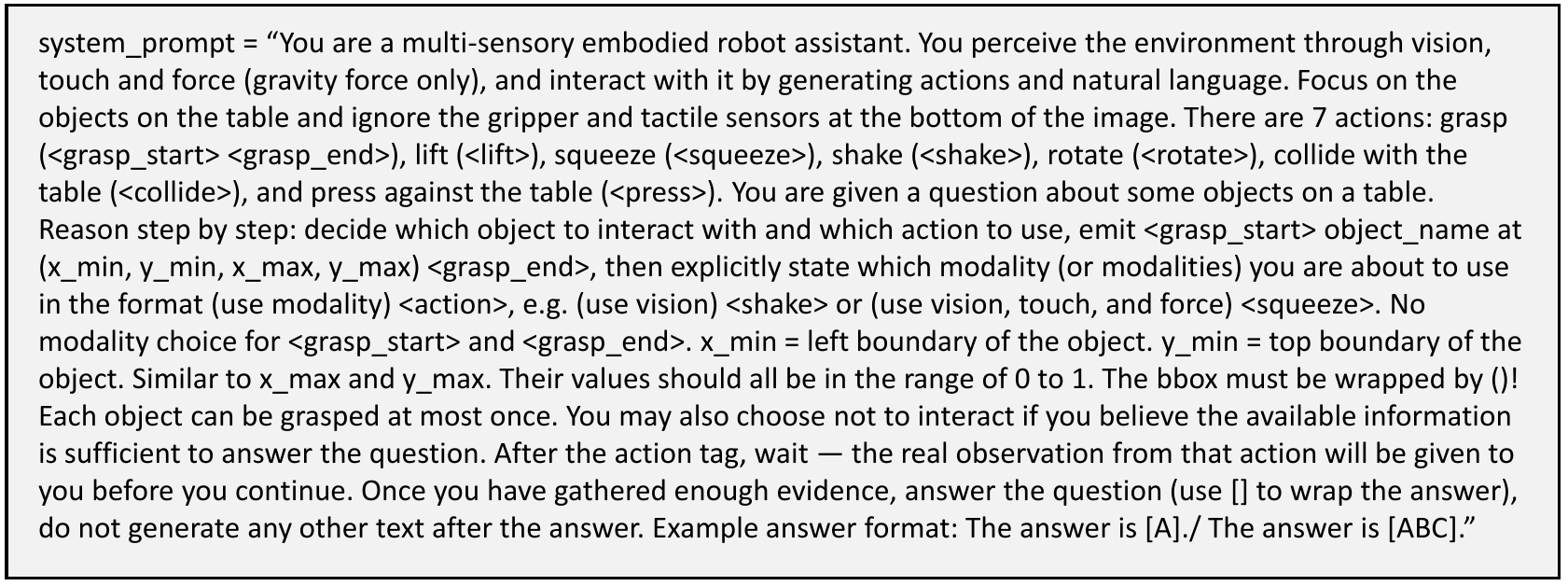}
    \caption{The system prompt for GPT-5.4 and GPT-6 Astra in ROMA Bench active perception evaluation.}
    \label{fig:gpt-sys}
\end{figure}

\begin{figure}[h!]
    \centering
    \includegraphics[width=0.9\linewidth]{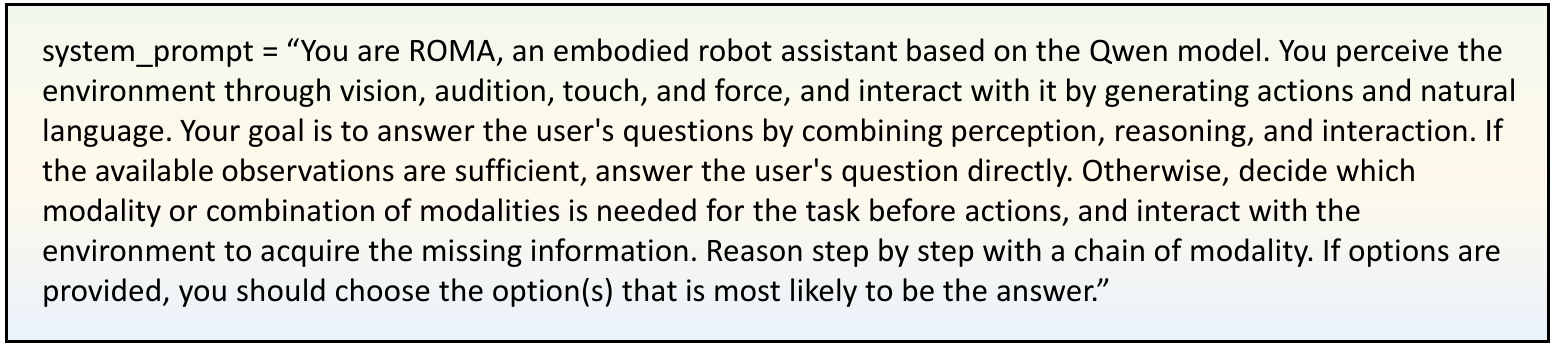}
    \caption{The system prompt for ROMA-7B in ROMA Bench active perception evaluation.}
    \label{fig:roma-sys}
\end{figure}

We initialize the ROMA-7B LLM from the weights of Qwen 2.5-Omni 7B~\citep{xu2025qwen25omnitechnicalreport} and incorporate the 2-frame version of the AnyTouch 2~\citep{feng2026anytouch2} tactile encoder. For stage 1 alignment, we construct modality-specific QA pairs from publicly available datasets and combine them with the handheld object subset of ROMI-2K. For audio alignment, we use AudioCaps~\citep{kim2019audiocaps} and AudioSet~\citep{gemmeke2017audio}. Specifically, we use AudioCaps and select AudioSet samples that contain object-related sounds to construct audio QA pairs. Since the resulting data pool is substantially larger than the audio data in ROMI-2K and has lower data quality, we randomly sample 300,000 QA pairs from them for stage 1 alignment. For tactile alignment, we use four publicly available datasets, Touch and Go~\citep{yang2022touch}, PHYSICLEAR~\citep{yu2024octopi}, TVL~\citep{fu2024touch}, and TacQuad~\citep{feng2025anytouch}, to construct tactile QA pairs. To avoid excessive redundancy caused by consecutive frames from the same contact event, we retain only one tactile sample for each contact, resulting in 10,821 QA pairs. We further combine these internet-scale audio and tactile QA pairs with the handheld object subset of ROMI-2K for stage 1 alignment.
For stage 2 SFT, we use both subsets of ROMI-2K and additionally mix in the HOTS tabletop object dataset~\citep{tziafas2026enhancing} during the early stage of training to improve tabletop object detection. The detailed parameter settings are shown in Tab.~\ref{tab:training-hyperparams}.

For offline evaluation, we use 2,100 active perception QA pairs from ROMA Bench, including multiple-choice, true-or-false, and ranking questions. For real-world evaluation, we randomly select 8 scenes from ROMA Bench and 132 questions from these scenes, and remove the answer choices to evaluate the model in a free-form question-answering setting. The system prompts used by Gemini 3.5 Flash, GPT-5.4, and ROMA-7B for evaluation on ROMA Bench are shown in Fig.~\ref{fig:gemini-sys}, \ref{fig:gpt-sys} and \ref{fig:roma-sys}.

Our baselines include both closed-source and open-source multi-modal LLMs. The closed-source baselines include GPT-5.4~\citep{openai2026gpt54} and Gemini 3.5 Flash~\citep{google2026gemini35}, while the open-source baselines include Qwen 2.5-Omni 7B and Qwen3-Omni 30B-A3B~\citep{xu2025qwen3}. For GPT-5.4, we use a reasoning effort of medium, while Gemini 3.5 Flash is evaluated with the low thinking level. Since each active perception task may require multiple model calls to iteratively select objects, choose interactions, and reason over newly acquired observations, using higher reasoning levels would substantially increase both inference latency and computational cost. 
We therefore adopt these settings to maintain a practical balance between reasoning capability, inference efficiency, and cost. 
For GPT-6 Astra, however, we set the reasoning effort to high to fully leverage its reasoning capability and provide a stronger reference for the performance achievable by current state-of-the-art MLLMs. This setting is intended to expose the model's upper-bound capability in active perception, rather than prioritize inference efficiency.
For all baselines, optical tactile images are provided as regular visual inputs. Since GPT-5.4 does not support object audio understanding in our evaluation setting, we provide the available non-audio modalities and require the model to reason from them instead. 


\section{Interaction and Ablation Study}
\label{app:ablation}

\begin{figure}[t]
    \centering
    \includegraphics[width=\linewidth]{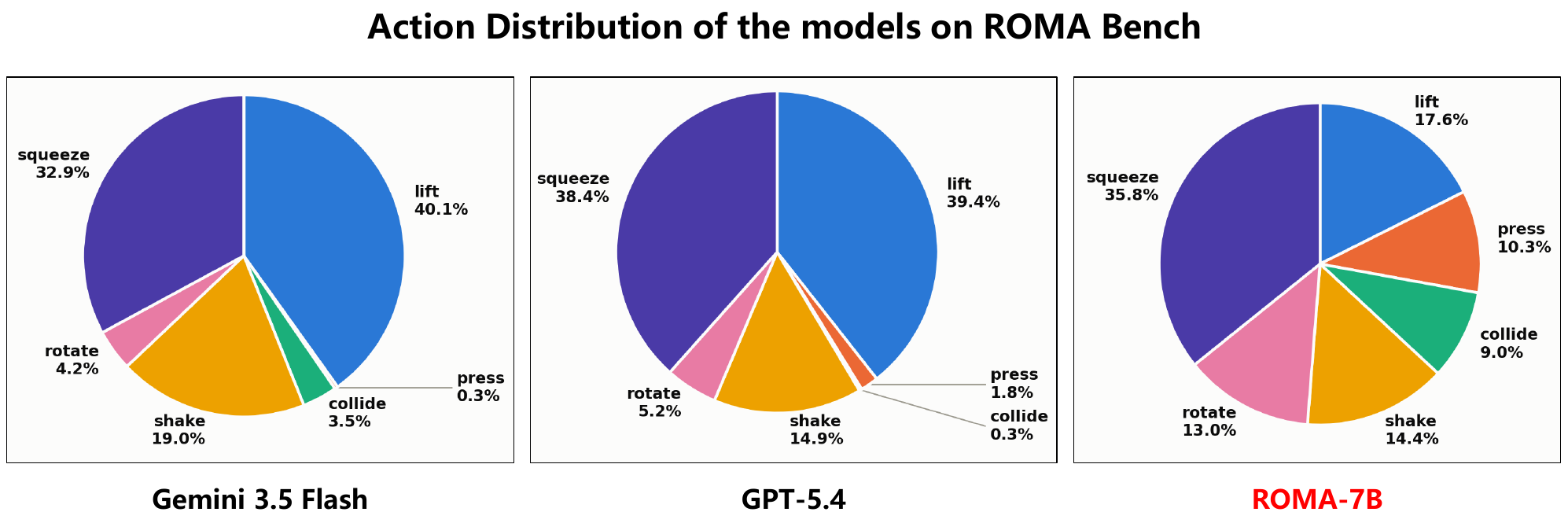}
    \caption{Comparison of action frequencies on completed active perception tasks in ROMA Bench.}
    \label{fig:actions}
\end{figure}

In this section, we first analyze the interaction patterns of different models on completed active perception tasks, followed by component-level ablations and missing-modality experiments.

\paragraph{Action-level Interaction Analysis.}
To analyze how different models explore the physical world and select interactions to acquire additional sensory evidence, we compare the relative distributions of the six interaction primitives performed by GPT-5.4, Gemini 3.5 Flash, and ROMA-7B on ROMA Bench, as shown in Fig.~\ref{fig:actions}. 
The results reveal clear differences in their interaction preferences. GPT-5.4's interaction distribution is dominated by the basic \textit{lift} and \textit{squeeze} operations, while \textit{collide}, \textit{shake}, \textit{rotate}, and \textit{press} account for smaller proportions. This interaction preference can limit the diversity of sensory evidence available for tasks involving material and object contents, where complementary interactions can provide additional relevant observations.
Gemini 3.5 Flash exhibits a broader interaction distribution than GPT-5.4, with larger proportions of \textit{shake}, \textit{rotate}, and \textit{collide}. Nevertheless, its interaction distribution remains concentrated on \textit{lift} and \textit{squeeze}, while the other actions occupy smaller proportions. Overall, its interaction preference remains similar to that of GPT-5.4.
In contrast, ROMA exhibits a more diverse interaction distribution and makes greater use of complementary interactions. In particular, \textit{rotate} complements \textit{shake} by providing additional audio evidence about object contents through a different motion pattern, while \textit{press} complements \textit{squeeze} by providing localized physical evidence for characterizing material, hardness, and surface properties. Meanwhile, \textit{collide} enables ROMA to probe the material properties of the object's bottom surface, which may not be sufficiently observable from the initial visual input. By combining these complementary interactions, ROMA can acquire more informative multi-sensory evidence for inferring the target attributes.
Given the relatively balanced attribute coverage of ROMA Bench shown in Fig.~\ref{fig:bench-xiantu}, ROMA's diverse interaction distribution further suggests that it can better assess whether the available sensory evidence is sufficient and selectively use complementary interactions to acquire task-relevant evidence across different attributes. 

\begin{table}[h]
\centering
\small
\caption{Evaluation of multi-sensory active perception on ROMA Bench under relaxed object localization criteria, with an emphasis on target object and interaction selection, as well as multi-sensory reasoning. Ha. (Hardness), Ro. (Roughness), Te. (Texture), In. (Inside), Ma. (Material), and We. (Weight) denote the target object attributes in single-chain and multi-chain perception tasks. We provide \textbf{oracle bounding boxes} among the answer options for single-chain and multi-chain tasks, and relax the bounding box matching criterion for all three task types. These settings largely reduce the impact of localization errors on physical interaction and subsequent multi-sensory reasoning, while retaining target object identification as part of the active perception process. It is important to note that providing oracle bounding boxes is \textbf{not a realistic setting} in real-world active perception. \textbf{Bold} and \underline{underlined} values indicate the best and second-best results, respectively.}
\label{tab:bbox_results}
\tabcolsep=0.065cm
\begin{tabular}{c|ccccccc|ccccccc|c|c}
\toprule
\multirow{2}{*}{\textbf{Model}}
& \multicolumn{7}{c|}{\textbf{Single-Chain}}
& \multicolumn{7}{c|}{\textbf{Multi-Chain}}
& \multirow{2}{*}{\makecell{\textbf{Intent-}\\\textbf{Driven}}}
& \multirow{2}{*}{\textbf{Total}} \\
\cmidrule(lr){2-8}
\cmidrule(lr){9-15}
& \textbf{Ha.} & \textbf{Ro.} & \textbf{Te.} & \textbf{In.} & \textbf{Ma.} & \textbf{We.} & \textbf{All}
& \textbf{Ha.} & \textbf{Ro.} & \textbf{Te.} & \textbf{In.} & \textbf{Ma.} & \textbf{We.} & \textbf{All}
& & \\
\midrule

GPT-5.4
& \underline{68.7} & 64.1 & 38.1 & 33.3 & 70.1 & 92.2 & 68.2
& 56.5 & 58.2 & 53.7 & 57.0 & 58.4 & 64.7 & 58.5
& 67.5 & 63.3 \\

GPT-6 Astra
& \textbf{73.7} & \textbf{94.9} & \underline{76.2} & 32.1 & \textbf{76.6} & \textbf{100.0} & \textbf{78.3}
& \underline{64.3} & \underline{73.3} & \underline{66.2} & 57.6 & 66.0 & \underline{75.5} & \underline{66.9}
& \textbf{83.9} & \underline{73.5} \\

Gemini 3.5 Flash
& 66.7 & \underline{82.1} & 73.8 & \underline{46.2} & 69.4 & \textbf{100.0} & 74.5
& 62.5 & 70.3 & 60.8 & \underline{61.7} & \underline{67.0} & 70.1 & 65.1
& 67.5 & 68.8 \\

Qwen 2.5-Omni
& 33.3 & 43.6 & 28.6 & 14.1 & 44.7 & 7.8 & 31.3
& 32.0 & 32.7 & 33.8 & 31.2 & 36.4 & 26.7 & 32.1
& 31.3 & 31.7 \\

Qwen 3-Omni
& 7.1 & 6.4 & 9.5 & 10.3 & 50.2 & 3.3 & 23.6
& 15.2 & 14.5 & 16.9 & 16.8 & 23.3 & 2.9 & 14.8
& 31.9 & 20.5 \\

\textbf{ROMA-7B}
& 54.5 & 75.6 & \textbf{83.3} & \textbf{66.7} & \underline{71.1} & \underline{97.4} & \underline{75.0}
& \textbf{66.5} & \textbf{79.4} & \textbf{79.6} & \textbf{77.4} & \textbf{73.0} & \textbf{80.9} & \textbf{76.4}
& \underline{77.1} & \textbf{76.0} \\

\bottomrule
\end{tabular}
\end{table}
\paragraph{Ablation of Object Localization Error Impact.}
To examine target object and interaction selection as well as multi-sensory reasoning while minimizing the impact of object localization errors, we evaluate the models under relaxed localization criteria, as shown in Tab.~\ref{tab:bbox_results}. For Single-Chain and Multi-Chain tasks, oracle bounding boxes are provided among the answer options, while the bounding box matching criterion is relaxed for all three task types. This reduces the effect of localization errors on physical interaction and subsequent multi-sensory reasoning while retaining target object identification as part of active perception.
This setting does not fully reflect real-world active perception and \textbf{should not be interpreted as an evaluation of complete active perception capability}, since object localization itself is an integral part of the process. Notably, GPT-6 Astra already exhibits very few localization errors under the original setting, and thus benefits only marginally from oracle bounding boxes.
Under this setting, GPT-6 Astra achieves the strongest performance among frontier MLLMs, particularly on Single-Chain tasks and visually accessible attributes such as hardness, roughness, material, and weight. GPT-5.4 and Gemini 3.5 Flash also remain competitive with ROMA-7B on these attributes, highlighting the strong visual and reasoning capabilities of frontier MLLMs. In contrast, ROMA-7B performs better on texture and inside-related attributes, where visual observations alone are often insufficient and complementary audio and tactile evidence are more informative.
This advantage becomes more pronounced on Multi-Chain tasks, where ROMA-7B achieves 76.4\%, substantially outperforming GPT-6 Astra (66.9\%), GPT-5.4 (58.5\%), and Gemini 3.5 Flash (65.1\%), and ranks first on nearly all attributes. Although GPT-6 Astra performs better on Intent-Driven tasks, ROMA-7B achieves the highest overall accuracy. These results indicate that ROMA's advantage extends beyond single-step attribute recognition to the active acquisition and integration of complementary sensory evidence over multiple interactions.
In comparison, the similarly sized Qwen 2.5-Omni and Qwen 3-Omni exhibit substantially more target object localization errors even with oracle bounding boxes, suggesting greater difficulty in following the benchmark's task-specific instructions. This further highlights the value of ROMI-2K, whose diverse real-world objects, physical interactions, and synchronized multi-sensory observations provide supervision for task-specific active perception. Building on this foundation, ROMA-7B achieves strong performance despite its 7B scale, demonstrating reliable instruction following and multi-sensory evidence integration.

\begin{table}[t]
\centering
\small
\caption{Ablation study of ROMA on ROMA Bench. Each row removes one component or training strategy from the full ROMA model.}
\label{tab:ablation}
\begin{tabular}{c|cccc}
\toprule
\textbf{Variant}
& \textbf{Single-Chain}
& \textbf{Multi-Chain}
& \textbf{Intent-Driven}
& \textbf{Total} \\
\midrule

\textbf{ROMA-7B}
& \textbf{71.1}
& \textbf{74.1}
& 73.1
& \textbf{72.9} \\

- Handheld Data
& 69.2
& 72.0
& 70.9
& 70.9 \\

- Token Init
& 69.0
& 73.3
& \textbf{73.4}
& 71.8 \\

- Dual Align
& 70.3
& 71.3
& 72.4
& 71.1 \\

- Sampling
& 70.0
& 73.1
& 70.0
& 71.5 \\

\bottomrule
\end{tabular}
\end{table}

\begin{table}[t]
\centering
\small
\caption{Evaluation of ROMA under missing-modality settings on ROMA Bench. Each variant returns ``Not Available'' when the model attempts to acquire the corresponding sensory modality.}
\label{tab:missing_modality}
\tabcolsep=0.065cm
\begin{tabular}{c|ccccccc|ccccccc|c|c}
\toprule
\multirow{2}{*}{\textbf{Variant}}
& \multicolumn{7}{c|}{\textbf{Single-Chain}}
& \multicolumn{7}{c|}{\textbf{Multi-Chain}}
& \multirow{2}{*}{\makecell{\textbf{Intent-}\\\textbf{Driven}}}
& \multirow{2}{*}{\textbf{Total}} \\
\cmidrule(lr){2-8}
\cmidrule(lr){9-15}
& \textbf{Ha.} & \textbf{Ro.} & \textbf{Te.} & \textbf{In.} & \textbf{Ma.} & \textbf{We.} & \textbf{All}
& \textbf{Ha.} & \textbf{Ro.} & \textbf{Te.} & \textbf{In.} & \textbf{Ma.} & \textbf{We.} & \textbf{All}
& & \\
\midrule

\textbf{ROMA-7B}
& \textbf{48.5} & \textbf{67.9} & \textbf{81.0} & 59.0 & \textbf{70.8} & \textbf{91.5} & \textbf{71.1}
& \textbf{62.5} & \textbf{75.8} & \textbf{77.4} & \textbf{75.1} & \textbf{70.7} & \textbf{80.9} & \textbf{74.1}
& \textbf{73.1} & \textbf{72.9} \\

- Audio
& 47.5 & \textbf{67.9} & \textbf{81.0} & 24.4 & 65.6 & \textbf{91.5} & 65.3
& 57.2 & 67.9 & 70.3 & 58.1 & 58.4 & 74.8 & 64.3
& 61.0 & 64.1 \\

- Touch
& 42.4 & 25.6 & 35.7 & 60.3 & \textbf{70.8} & \textbf{91.5} & 63.4
& 60.6 & 29.7 & 43.1 & 64.6 & 67.5 & 72.5 & 59.5
& 71.5 & 62.7 \\

- Force
& \textbf{48.5} & \textbf{67.9} & \textbf{81.0} & \textbf{61.5} & 70.4 & 46.4 & 61.9
& 57.2 & 70.9 & 67.3 & 64.3 & 61.8 & 52.7 & 61.6
& 65.0 & 62.2 \\

\bottomrule
\end{tabular}
\end{table}
 
\paragraph{Component and training ablation.}
To analyze the contribution of different components and training strategies in ROMA, we evaluate a series of ablated variants, as shown in Table~\ref{tab:ablation}. The full ROMA model achieves an overall accuracy of 72.9\%, while removing any individual component or training strategy leads to a performance drop. These results demonstrate that the final performance of ROMA benefits from the joint contribution of its multi-sensory architecture, interaction data, modality alignment, and training strategy. The handheld-data ablation further highlights the importance of diverse object-level interaction trajectories. Although these data are collected outside the final tabletop setting, they provide broader coverage of object identities, grasp configurations, and interaction outcomes, improving the model's ability to generalize to diverse objects and physical interactions. Moreover, removing our special token initialization strategy also degrades performance, suggesting that appropriate initialization facilitates the integration of the newly introduced action and modality tokens into the pre-trained multi-modal model. 
The dual alignment ablation, which removes the audio-visual-text and tactile-visual-text alignments while retaining the conventional audio-text and tactile-text alignments, results in a performance drop. This demonstrates the importance of jointly aligning sensory observations with both visual and linguistic representations for effective cross-modal reasoning.
Finally, removing the dynamic scene sampling strategy also reduces performance. This indicates that simply combining diverse interaction data is insufficient, and that appropriately balancing different data sources is important for effective training. The sampling strategy helps progressively shift the training distribution from object-level interaction data toward real-world scene-level active perception, enabling the model to better adapt its learned multi-sensory interaction knowledge to the active perception task.

\paragraph{Missing-modality analysis.}
To examine the contribution of different sensory modalities and the stability of ROMA when sensory information is unavailable, we evaluate ROMA under missing-modality settings in Tab.~\ref{tab:missing_modality}. When ROMA attempts to acquire an unavailable modality, the system returns ``Not Available'' instead of providing the corresponding observation, while the remaining interaction and reasoning process is kept unchanged. The results show that removing audio leads to a clear degradation on tasks involving object contents, where \textit{shake} and \textit{rotate} provide informative interaction-induced sounds. Removing tactile information mainly affects tasks requiring local contact evidence, such as roughness, hardness, and material properties. Similarly, removing force information degrades performance on tasks that rely on force-related evidence, especially weight-related cues. These results demonstrate that audio, tactile, and force sensing provide complementary information for active perception. However, we observe an unexpected phenomenon: removing the force modality improves performance on tasks involving object contents, where performance would normally be expected to degrade. This suggests that ROMA still has room to improve in effectively leveraging force for the complex reasoning required to infer object contents. Fully integrating force sensing into embodied multi-sensory foundation models and agents remains an open challenge. Meanwhile, the missing-modality setting evaluates whether ROMA can maintain stable perception and reasoning when part of the expected sensory evidence is unavailable. Although performance degrades when a modality is missing, the tasks do not completely fail, indicating that ROMA can adapt to incomplete sensory evidence and maintain its perception capability rather than collapsing when one of the sensory modalities is unavailable.

\end{document}